\documentclass[11pt]{article}
\usepackage[utf8]{inputenc}
\usepackage{textcomp}
\usepackage[
  left=2.5cm,
  right=2.5cm,
  top=3cm,
  bottom=3cm
]{geometry}
\usepackage{setspace}
\usepackage{titlesec}
\usepackage{graphicx}
\usepackage{siunitx}
\usepackage{longtable,tabularx}
\usepackage{amsmath, amsfonts, bm}
\usepackage{graphicx}
\usepackage{rotating, threeparttable, multirow, multicol, makecell, booktabs}
\usepackage{array}
\usepackage{indentfirst}
\usepackage{subcaption}
\usepackage{pgfplots}
\usepackage{hyperref}
\tikzset{every picture/.style={font=\normalsize}}
\newcolumntype{M}[1]{>{\centering\arraybackslash}m{#1}} 
\usepackage{float}
\usepackage{authblk}

\usepackage{soul} 
\newif\ifhighlight
\highlightfalse  
\sethlcolor{yellow!25}
\newcommand{\modified}[1]{%
  \ifhighlight
    \hl{#1}%
  \else
    #1%
  \fi
}

\usepackage{tcolorbox}
\tcbuselibrary{skins, breakable}
\newenvironment{modifiedblock}
{%
  \ifhighlight
    \begin{tcolorbox}[
      colback=yellow!25,
      colframe=yellow!25,
      boxrule=0pt,
      arc=2pt,
      left=4pt,
      right=4pt,
      top=4pt,
      bottom=4pt,
      breakable
    ]
  \else
    \begingroup
  \fi
}
{%
  \ifhighlight
    \end{tcolorbox}
  \else
    \endgroup
  \fi
}

\usepackage[table]{xcolor}
\newcommand{\mdf}{%
  \ifhighlight
    \cellcolor{yellow!25}%
  \fi
}

\title{\modified{Towards a Foundation-Model Paradigm for Aerodynamic Prediction in Three-dimensional Design}}

\author[1]{Yunjia Yang \footnote{Postdoctoral researcher, School of Computation, Information and Technology, Email: yunjia.yang@tum.de, Corresponding author}}
\affil[1]{Technical University of Munich, Garching, D-85748, Germany}
\author[2]{Babak Gholami\footnote{Product Owner, Generative Design and AI for CAx, Email: Babak.Gholami@bmw.de}}
\author[2]{Caglar Gurbuz\footnote{Systems Engineer, Generative Design and AI for CAx, Email: Caglar.Gurbuz@bmw.de}}
\author[2,1]{Mohammad Rashed\footnote{Ph.D. candidate, Generative Design and AI for CAx, Email: m.rashed@tum.de}}
\affil[2]{BMW Group, Bremer Str. 6, Munich, D-80788, Germany}
\author[1]{Nils Thuerey\footnote{Professor, School of Computation, Information and Technology, Email: nils.thuerey@tum.de}}
\date{}
\begin{document}

\maketitle

\begin{abstract}
\noindent\normalsize 
Accurate machine-learning models for aerodynamic prediction are essential for accelerating shape optimization, yet remain challenging to develop for complex three-dimensional configurations due to the high cost of generating training data. This work introduces a methodology for efficiently constructing accurate surrogate models for design purposes by first pre-training a large-scale model on diverse geometries and then fine-tuning it with a few more detailed task-specific samples. A Transformer-based architecture, AeroTransformer, is developed and tailored for large-scale training to learn aerodynamics. The methodology is evaluated on transonic wings, where the model is pre-trained on SuperWing, a dataset of nearly 30000 samples with broad geometric diversity, and subsequently fine-tuned to handle specific wing shapes perturbed from the Common Research Model. \modified{Results show that, with 450 task-specific samples, the proposed methodology achieves 0.36\% error on surface-flow prediction, reducing 84.2\% compared to training from scratch. The influence of model configurations and training strategies is also systematically studied to provide guidance on effectively training and deploying such models under limited data and computational budgets. To facilitate reuse, we release the datasets and the pre-trained models at }\href{https://github.com/tum-pbs/AeroTransformer}{https://github.com/tum-pbs/AeroTransformer}. An interactive design tool is also built on the pre-trained model and is available online at \href{https://webwing.pbs.cit.tum.de}{https://webwing.pbs.cit.tum.de}.

\end{abstract}

\newpage

\section{Introduction}

Fast predictions of aerodynamic coefficients and flow fields around three-dimensional components, such as transonic swept wings, are essential for accelerating aerodynamic design in the industry \cite{brunton_data-driven_2021}. Although computational fluid dynamics (CFD)-based optimization methods, particularly those utilizing adjoint techniques for gradient evaluation, have been applied for efficient shape optimization, they remain computationally expensive~\cite{martins_aerodynamic_2022}. This is especially true for multipoint and robust optimization tasks, which require repeated aerodynamic evaluations across various design points \cite{yang_fast_2024, ma_robust_adjoint_2023, kenway_multipoint_2016}.

Machine-learning techniques have emerged in this decade as a promising tool to reshape the optimization process for aerodynamic shape design~\cite{li_machine_2022}. Pre-trained surrogate models~\cite{thuerey2020dfp,chen2021highacc} have been extensively studied to accelerate the evaluation of aerodynamic performance during optimization, but their practical deployment is often limited by the pre-training cost~\cite{renganathan_enhanced_2021, iliadis_dnn-driven_2023}. The surrogate models in most applications are relatively local to a \textit{specific optimization problem domain}, trained only with samples similar to the baseline shape \cite{li_dbo_2021,chen2021numerical,hasan_wing_bspline_spanwise_crm_2025}. However, this is often not economical, since generating a sufficiently rich dataset with high-fidelity CFD can easily cost more than the savings from subsequent optimization runs using the trained model. Therefore, a more viable strategy is to construct surrogate models that are flexible, and as general as possible, so that the same model and its variants can be reused across many optimization tasks in the \textit{design space}.

Existing work has demonstrated that such \textit{general} models are feasible in two-dimensional settings, particularly for airfoils \cite{yang_fast_2024, li_dbo_2019, bouhlel_scalable_airfoil_2020, renganathan_enhanced_2021, tian_dbo_pressureguide_2024, yang_generalizable_2025}. However, extending this approach to three-dimensional components such as automobile or aircraft wings introduces new challenges. On the one hand, the design space becomes extremely high-dimensional, especially when realistic details are considered \cite{wu_sensitivity-based_2024, chen_aerodynamic_2025}. Sampling this space using physics-based simulations induces a high computational cost. On the other hand, design optimization algorithms are highly sensitive to the predicted performance, and even modest errors can mislead the optimizer and cause divergence \cite{yang_uadbo}, necessitating high accuracy within the problem domain. This challenge is demonstrated by the existing datasets for vehicles and aircraft: They either contain many samples but rely on simplified geometries / limited design spaces (e.g. ShapeNet Cars \cite{umetani_learning_2018}, Emmi-Wing \cite{paischer_going_2025}, and CRM wing dataset \cite{li_dbo_2021}); or they provide detailed, industrial shapes but only in small numbers (e.g. DriveAerML Cars \cite{ashton_drivaerml_2025}, and the CRM WBPN dataset \cite{peter_oneras_2025}). Consequently, constructing a surrogate that covers a general design space and simultaneously meets the demanding accuracy requirements is difficult under realistic computational constraints.

In parallel, \textit{foundation models} have emerged as a new paradigm in machine learning
. By pre-training large models on vast and diverse datasets, they learn general-purpose representations that can be efficiently adapted to a wide range of downstream tasks with relatively little task-specific data. This paradigm has achieved striking success in natural language processing and has recently influenced scientific machine learning as well. Several studies have proposed large-scale scientific foundation models that generalize across different partial differential equations (PDEs) and initial and boundary conditions \cite{wu_transolver_2024, herde_poseidon_2024, xu_self-trans_2024, holzschuh_pde-transformer_2025, zhou_unisolver_2025, luo_mmet_2025}. However, most existing scientific foundation models focus primarily on basic flow within regular geometric domains, whereas, from the perspective of shape optimization, a foundation model that can be generalized across different shapes is desired. While some prior studies explore transferring pre-trained aerodynamic models to downstream tasks \cite{shen_geometric-perspective_2024, zhang_deep_2025, lin_transferable_2026}, their pre-training datasets are typically limited in scale and diversity. This restricts the model’s ability to learn broadly transferable representations, often resulting in limited downstream performance.

\modified{
In this paper, the aerodynamic surrogate modeling for shape optimization is viewed through the lens of the foundation-model paradigm, aiming to balance reusability and accuracy within an acceptable computational cost for building aerodynamic surrogates of three-dimensional shapes. We note one important aspect that distinguishes the use of a large language foundation model from that aimed at aerodynamic prediction: the high data-generation costs. To this end, we propose a strategy that balances the fidelity of geometric parameterization between pre-training and fine-tuning datasets. Based on this framework, we develop a scalable transformer-based surrogate model and systematically evaluate the proposed paradigm on large-scale wing datasets. Through extensive experiments, we demonstrate that large-scale pre-training significantly improves performance on downstream aerodynamic prediction tasks, which validates key properties expected of foundation-style models. We also examine different pre-training and fine-tuning strategies, and provide first evidence on how to efficiently use a pre-trained model for downstream tasks.
}

\section{\modified{The foundation-model paradigm for aerodynamic problems}}\label{sec:paradigm}

\begin{figure}[ht]
    \centering
    \includegraphics[width=0.8\linewidth]{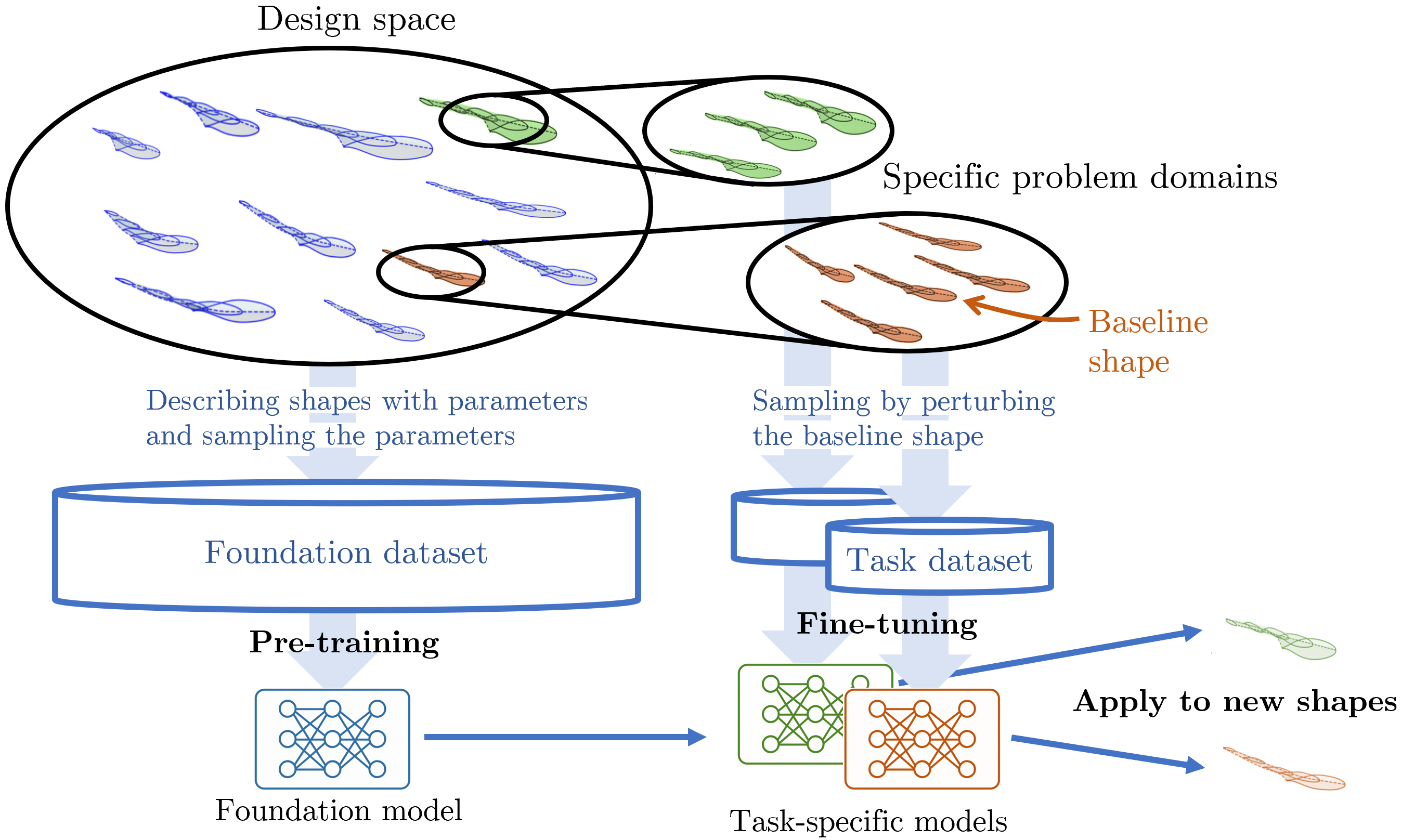}
    \caption{
    Our approach to building and utilizing foundation-model paradigm for aerodynamic design}
    \label{fig:framework}
\end{figure}

\modified{
The foundation-model paradigm relies on large, diverse datasets to learn broadly transferable relationships that can generalize across tasks. This principle motivates our approach to aerodynamic surrogate modeling, in which a model is first pre-trained on a broad dataset and later fine-tuned for specific design tasks.

However, applying this paradigm to aerodynamic analysis presents a fundamental challenge. Time-consuming CFD simulations make it prohibitively expensive to construct datasets that are both large-scale and rich in geometric detail. To address this challenge, we introduce a design principle to balance the fidelity of shape parameterization between the pre-training and fine-tuning stages. As illustrated in Fig.}~\ref{fig:framework} \modified{, we assign different requirements for geometric diversity and geometric detail across the two stages of the learning process.

In the pre-training stage, diversity is given priority to enable the model to learn dominant flow physics and the relationships between geometry and flow. The geometric details, on the other hand, are intentionally simplified to keep computational cost tractable.} Such parameterizations are commonly derived from engineering design practices, including Class–Shape Transformation (CST) \cite{du_modification_2024} and spline-based representations \cite{hasan_wing_bspline_spanwise_crm_2025}, or from data-driven approaches that extract dominant shape modes \cite{li_dbo_2021, xie_parametric_2024}. These reduced representations enable efficient sampling of a wide variety of geometries while keeping the computational cost manageable.

In the fine-tuning stage, we therefore restore geometric fidelity within a localized design domain, typically defined around a baseline configuration associated with a specific design task. Because this domain is restricted, richer parameterizations can be introduced without overwhelming computational cost. This can be achieved, for example, by perturbing the baseline geometry using deformation techniques such as Free Form Deformation (FFD) or by introducing higher-resolution geometric parameters.

\modified{
By restoring geometric detail only in the task-specific fine-tuning stage, the pre-trained base model can be efficiently adapted into an accurate local surrogate for downstream aerodynamic design problems. In this way, the proposed framework makes the foundation-model paradigm feasible for aerodynamic problems.
} 

\section{Scalable Transformer architecture and training strategies}

Transformer architectures have become the dominant backbone for contemporary foundation models. Their success stems from the self-attention mechanism, which enables flexible modeling of long-range dependencies, making them a natural candidate for learning generalizable representations of complex physical systems. To apply this paradigm to aerodynamic performance modeling, we introduce several adaptations to develop the \textit{AeroTransformer}, which is based on the PDE-Tranformer \cite{holzschuh_pde-transformer_2025}. We integrate this architectural choice with a tailored training and fine-tuning strategy that enables the model to adapt to task-specific tasks. 

In the following sections, we first briefly recall the fundamentals of the Transformer, Vision Transformer, and the PDE-Transformer. Then, we provide a detailed description of the AeroTransformer architecture, followed by the pre-training workflow and the fine-tuning methodology.

\subsection{Basics of Transformer Architectures} 

\subsubsection{Transformer and attention mechanism}\label{sec:transformer}

\begin{figure}[htbp]
    \centering
    \includegraphics[width=0.45\linewidth]{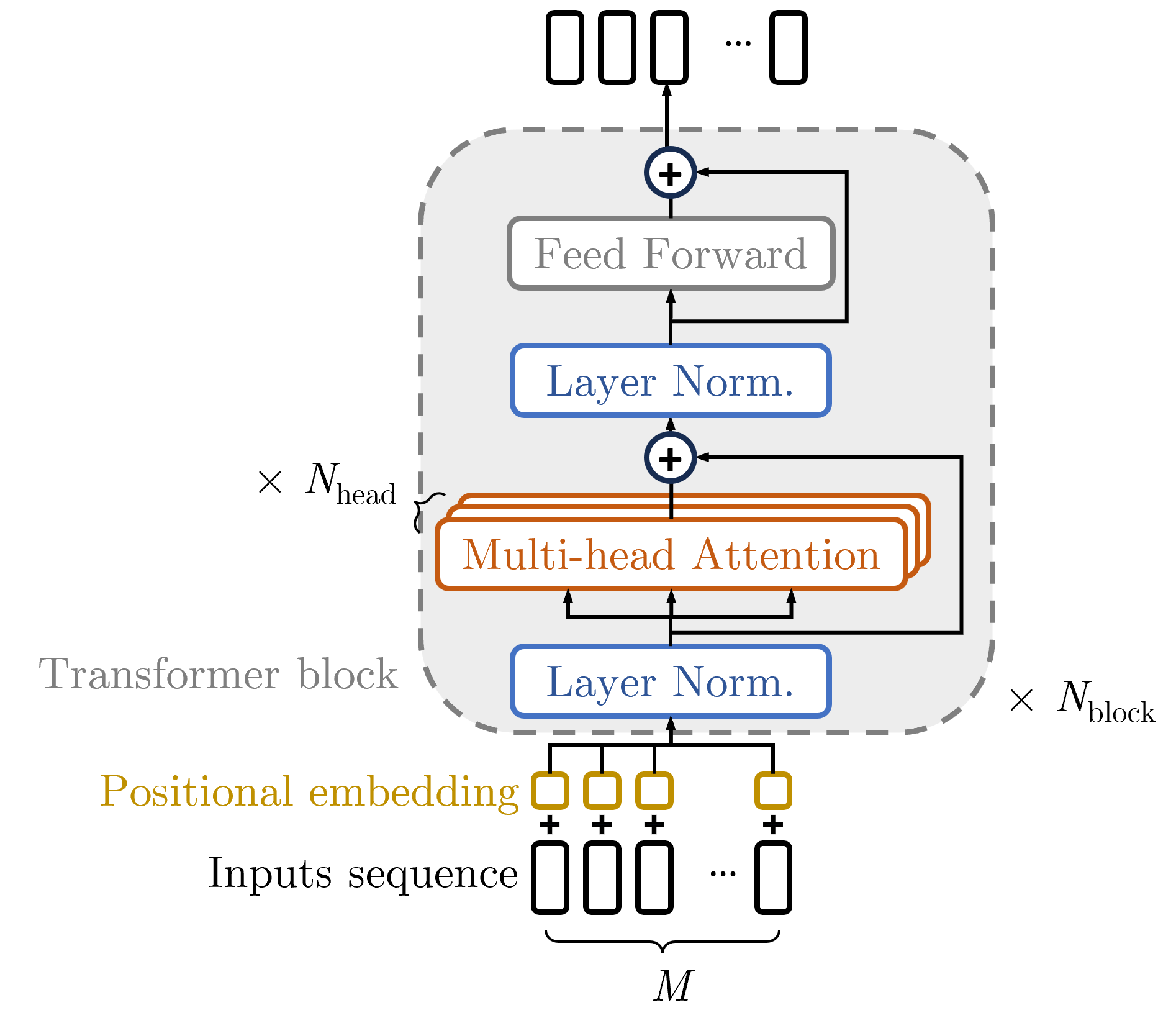}
    \caption{Transformer architecture}
    \label{fig:trans}
\end{figure}

The typical Transformer~\cite{vaswani_attention_2017} is an encoder-only architecture consisting of a stack of Transformer blocks. These blocks all share the same architecture, featuring a multi-head self-attention mechanism layer and a position-wise feedforward layer, both of which are followed by residual connections and layer normalization, as shown in Fig. \ref{fig:trans}.

The self-attention mechanism represents the core of the architecture. It determines how much each token should focus on every other token in the input sequence and helps the model learns the global correlation between every token. It is realized with a query-answer mechanism. Suppose the input sequence has $M$ tokens with each having $N_\mathrm{hidden}$ dimensions, a value vector $\bm v_i$ with dimension of $D$ is calculated for each token with shared weights, and formulate a value matrix $\bm V$ of size $M \times D$. Meanwhile, the query and key vectors $\bm{q}_i, \bm{k}_i$ are also calculated, contributing to the query and key matrices $\bm{Q}$ and $\bm{K}$. Then, for each token position $j = 0, \cdots, M$, the query vector $\bm{q}_j$ is compared with the key vector of every token position $\{\bm{k}_i\}_{i=0,\cdots M}$ to get the similarity (described most commonly by the scaled dot-product) between $\bm{q}_j$ and the key vectors. Finally, the output at the token position $j$ is calculated by summing up the value vectors $\bm{v}_i$ with the similarity as weights. The process can be written as:

\begin{modifiedblock}
\begin{equation}
    Att_j = \sum_i\mathrm{Softmax}_i\left(\frac{1}{\sqrt{D}}\bm{q}_j^T \cdot \bm{k}_i\right)\bm{v}_i
\end{equation} 
\end{modifiedblock}
where Softmax is used to normalize the attention weights, and $\sqrt D$ is the embedding dimension, used as a scaling factor to prevent large dot products. This mechanism enables each token to interact with every other token in the sequence, thereby enhancing the model’s ability to capture global dependencies. Meanwhile, it is highly parallelized, significantly speeding up training and inference.


Since a Transformer treats all input tokens equally, fixed or trainable positional embeddings are added to each input token in the sequence before feeding it into the encoder, in order to preserve positional information. The most common fixed positional embedding is the sinusoidal one; however, a trainable positional embedding offers more flexibility, which can result in improved results. To further enhance expressivity, many Transformers use multi-head attention, which projects the queries, keys, and values into multiple subspaces and performs parallel attention operations. The outputs of all heads are then concatenated and linearly transformed. This allows the model to jointly capture information from different representation subspaces. 


While the Transformer architecture was originally designed for sequential data, such as text, its success inspired researchers to explore its application to computer vision tasks. 
%
The ViT \cite{dosovitskiy_vit_2022} extends the Transformer from natural language processing to computer vision tasks. It divides an input image of size $H \times W$ into a sequence of non-overlapping fixed-size patches, each of size $p_H \times p_W$. Each patch is then flattened into a vector of size $p_H \times p_W$ and linearly projected into a lower-dimensional embedding space of dimension $N_\mathrm{hidden}$ using a learnable matrix. This yields a sequence of patch embeddings with a length of $M = (H \times W) / (p_H \times p_W)$, which serve as input tokens to the Transformer. 

\subsubsection{PDE-Transformer}

An extended variant of Vision Transformers, the PDE-Transformer \cite{holzschuh_pde-transformer_2025}, was recently proposed for PDE inference tasks, integrating several newly developed techniques to enhance performance and scalability. The PDE-Transformer deals with the spatiotemporal system $S$ that encompasses $n$ physical quantities $u(x,t):\Omega_S \times [0, T] \to \mathbb{R}^n$. Specifically, it predicts the snapshots in a discretized time sequence $\mathbf{u}_0^S, \mathbf{u}_{\Delta t}^S, \cdots, \mathbf{u}_T^S$, where the physical quantities on each snapshot are spatially discretized with a structured mesh $\mathbf{u}^S: \mathbb{R}^{H \times W}$. PDE-Transformer also adopts an autoregressive approach for prediction, i.e., it learns the mapping $\left[\mathbf{u}^S_{t-T_p\Delta t}, \cdots, \mathbf{u}^S_{t-\Delta t}\right] \to \mathbf{u}^S_t$.
As this architecture represents the basis for the proposed architecture, we summarize its key features below. 
We also provide a visual overview of PDE-Transformer in Fig. \ref{fig:aerotransframe}.

\paragraph{Input patching} The input of PDE-Transformer is the discretized physical quantities in the previous time step. PDE-Transformer first embeds the input into tokens using a convolutional embedding layer, preserving the input's dimensionality. Specifically, the input is partitioned into non-overlapping patches of size $p_H \times p_W$, and then each patch is mapped into a token with $N_{\mathrm{hidden},0}$ dimensions using a shared kernel. The output of sizes $(H/p_H)\times(W/p_W)\times N_{\mathrm{hidden},0}$ can be used for upcoming attention calculations.

\paragraph{Hierarchical architecture} The resolution of the flow field strongly influences the accuracy in capturing the key flow structures (i.e., the shock waves) and in reconstructing crucial aerodynamic coefficients. However, high resolutions result in a large number of tokens, limiting the model's scalability. To overcome this, a hierarchical U-shaped architecture is adopted. PDE-Transformer comprises several Stages, each containing several Transformer blocks. In the first half of the architecture, a down-sampling module is added at the end of the Transformer blocks to reduce the resolution of $H$ and $W$ by a factor of 2 while doubling the hidden dimension. \modified{This fixed $2\times2$ spatial reduction follows the common hierarchical design, allowing the model to gradually aggregate information.} In the middle, a stage without up- or down-sampling is used for the latent process. The last half of the stages has an up-sampling module before the first block in the stage to multiply the resolution and halve the hidden dimension. Skip connections are built between corresponding stages to inject the token features during down-sampling, thereby reintegrating them into the up-sampling path. Down- and up-sampling is performed using convolutional layers with PixelUnshuffle and PixelShuffle operations. 

\paragraph{Shifted Windowed attention} Another computational burden comes from the global attention calculation of ViT. PDE-Transformer utilizes the windowed multi-head self-attention (W-MSA) and shifted \modified{window} multi-head self-attention (SW-MSA) proposed in Swin-Transformer \cite{liu_swin_2021}. The global tokens are partitioned into windows of size $w \times w$, and the attention calculation is limited inside these local windows. To avoid discontinuities, the windows are shifted by $w/2$ tokens between two adjacent blocks with a mask. The positional embedding of the tokens is realized by combining the log-spaced relative positions of tokens with a feed-forward neural network.

\subsection{AeroTransformer}

The AeroTransformer is developed based on the PDE-Transformer architectures outlined above and is specifically modified for aerodynamic prediction tasks: to predict the aerodynamic performance given the geometry $\bm g$ and operating condition $\bm c$. In this paper, we focus on two-dimensional surface geometries in three-dimensional space, which can be mapping to a single-block structured mesh, i.e., $\bm g \in \mathbb{R}^{H \times W \times 3}$ (an example of the mapping can be found in the \textit{Experiments} section). The operating condition is a low-dimensional vector $\bm c \in \mathbb{R}^{N_c}$.

\begin{figure}[htbp]
    \centering
    \includegraphics[width=1\linewidth]{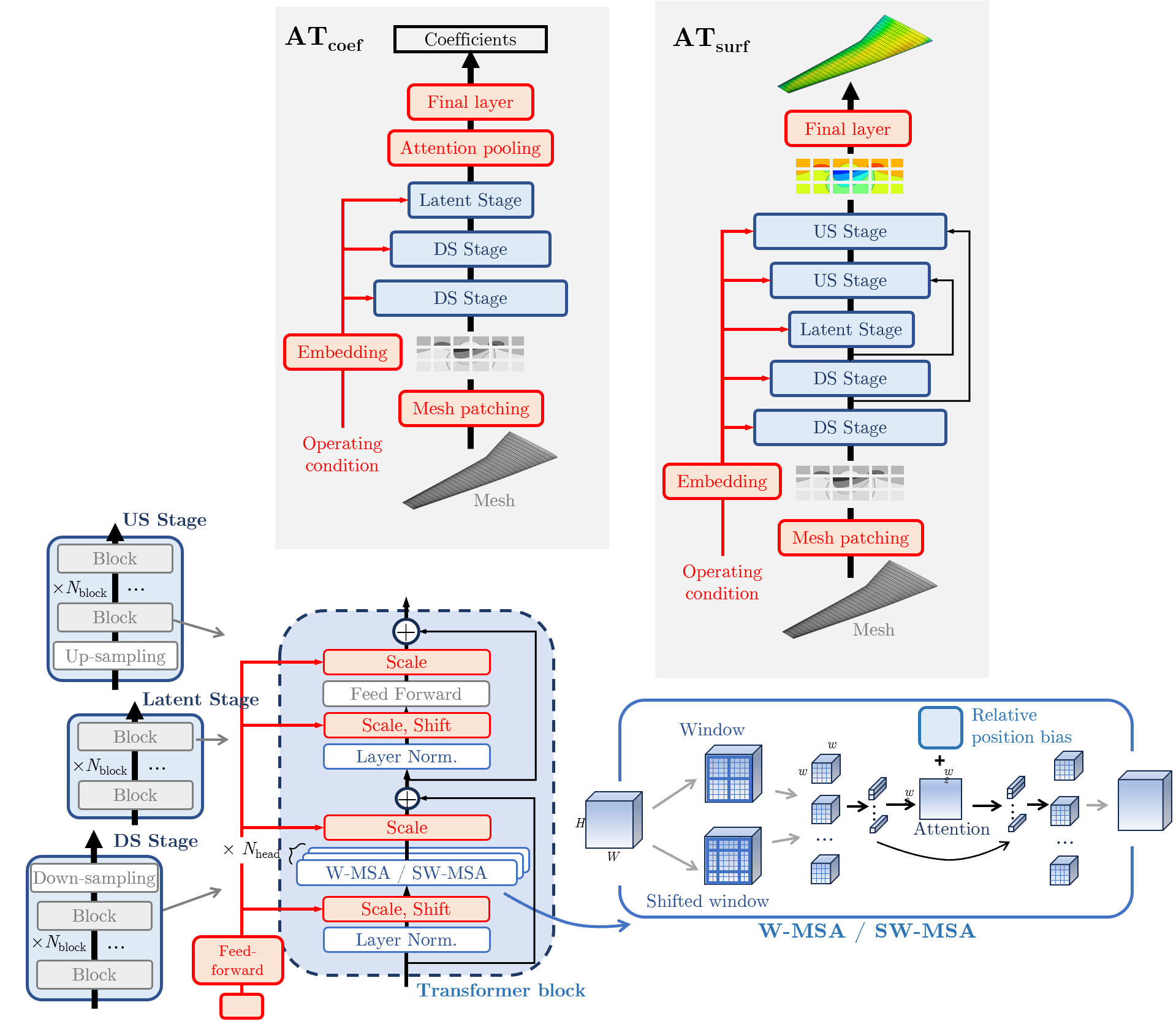}
    \caption{A visual overview of the AeroTransformer architecture}
    \label{fig:aerotransframe}
\end{figure}

From the perspective of design optimization, both the aerodynamic coefficients (such as the drag coefficient $C_D$) and the flow fields (especially the surface flow, as represented by the surface pressure coefficients $C_p$) are important. Although the coefficients are the primary quantities to drive the optimization, the flow fields can provide intuitive and physics-based information that is crucial for practical design solutions. Consequently, we provide two variants of AeroTransformer that can be employed for both kinds of inference tasks. The first surface flow AeroTransformer ($\mathrm{AT}_\mathrm{surf}$) predicts the surface flow field with $N_\mathrm{var}$ flow quantities $\mathbf{u}\in \mathbb{R}^{H\times W \times N_\mathrm{var}}$ on the input surface geometry $\bm g$, and the second coefficient AeroTransformer ($\mathrm{AT}_\mathrm{coef}$) directly predict the $N_k$ aerodynamic coefficients $\bm k \in \mathbb{R}^{N_k}$. Based on this task, we primarily modified the input and output of the PDE-Transformer, as detailed below. A sketch of the AeroTransformer architecture is provided in Fig. \ref{fig:aerotransframe}, where the PDE-Transformer backbone is shown with a blue background, and our modifications to enable aerodynamic prediction tasks are colored red.

\subsubsection{Shape and operating condition inputs}\label{sec:aeroinput}

\paragraph{Mesh patching} The mesh, instead of the previous flow field, acts as the primary input for AeroTransformer. It has the exact same shape as the flow field, enabling us to retain the same embedding layer of PDE-Transformers.

\paragraph{Operating condition injection} Besides the wing shapes, the operating conditions are an important input for the model. A trivial implementation for inputting operating conditions involves expanding them to the size of mesh points and concatenating them together for mesh embedding. However, this may lead to forgetting the operating conditions in later processes. In AeroTransformer, an operating condition injection approach is adopted, inspired by the adaptive layer normalization-zero (adaLN-Zero) conditioning mechanism from Diffusion Transformer \cite{peebles_scalable_2023}. The operating conditions first pass a small MLP to get the operating condition embedding vector. For every Transformer block, the tokens after layer normalization are shifted and scaled before being fed into the next layer. The manipulated tokens after the attention and feed-forward layers are also scaled. The scale and shift vectors are regressed from the operating condition embedding vector using a feed-forward layer for each block. 

\subsubsection{Aerodynamic outputs and loss terms}\label{sec:aerooutput}

As mentioned earlier, AeroTransformer is designed for both surface flow and coefficient prediction tasks, as already shown in Fig. \ref{fig:aerotransframe}. Here, we introduce the architectures of both models and present an additional method for integrating surface flow output into the coefficients. 

\paragraph{$\mathbf{AT}_\mathbf{surf}$}

The $\mathrm{AT}_\mathrm{surf}$ model produces a high-dimensional output: the surface flow field at the same mesh-centric points as the input shape mesh. To reconstruct the surface flow, a final convolutional layer is added. It expands the token dimension to $N_{\mathrm{var}}\times p_H \times p_W$, where $N_{\mathrm{var}}$ is the number of channels of the output flow, and $p_H, p_W$ are the patching size in the mesh embedding. Then, these tokens can be easily reshaped for the surface flow output.

\paragraph{$\mathbf{AT}_\mathbf{coef}$}

The $\mathrm{AT}_\mathrm{coeff}$ model directly outputs the aerodynamic coefficients $\bm k$ with an Encoder-only architecture: Only the first half and the latent stages of the original are used, and the latent tokens with shape $N_{\mathrm{hidden}, l} \times H_l \times W_l$ are further compressed with an attention-based pooling layer to a vector of dimension $N_{\mathrm{hidden}, l}$. This vector is then mapped to the output coefficients with a fully connected layer.

\paragraph{Aerodynamic coefficients with $\mathbf{AT}_\mathbf{surf}$}

$\mathrm{AT}_\mathrm{coef}$ provides a direct way for modeling aerodynamic coefficients, but previous research \cite{yang_fast_2024} has shown on two-dimensional airfoils that this approach can suffer from generalization issues. 
To address this challenge, we also adopt an alternative approach, using the $\mathrm{AT}_\mathrm{surf}$ model and obtaining the coefficients by integration, since the primary surface flow output contains sufficient information. For example, the lift coefficient ($C_L$) and drag coefficients ($C_D$) can be integrated from the surface pressure and friction coefficients as follows:

\begin{align}\label{eqn:force}
    &\left[C_L, C_D\right] = \bm R_\alpha \cdot C_{\bm F} \nonumber\\
    &C_{\bm{F}}=\sum_{i=1}^{N_{\mathrm{cell}}}C_{p,i}\cdot\bm n_i \cdot A_i + (C_{\bm f, i} - C_{\bm f, i} \cdot \bm n_i) \cdot A_i 
\end{align}
where $R_\alpha$ is the rotation matrix for angle of attack $\alpha$, $n_i$ is the normal vector of the surface mesh cell, and $A_i$ is the area of it.

\paragraph{Extra loss term for $\mathbf{AT}_\mathbf{surf}$} \label{sec:losses}

In the second approach, where coefficients are obtained from surface flow, we expect that the accuracy of the integrated coefficients should rely on the accuracy of the surface flow. However, this is not always true because both approaches use different ways to calculate training losses. For the surface flow, we use the mean square error (MSE) for the loss function:

\begin{align}
    \mathcal L_{\mathrm{surf}} &= \mathrm{MSE}\left(\hat C_p, \hat C_{\bm f};  C_p, C_{\bm f}\right)\nonumber\\
    &=\sum_{i=1}^{N_\mathrm{cell}}\frac{1}{N_\mathrm{cell}}\left[\left(\hat C_{p,i}-C_{p,i}\right)^2 + \left\Vert \hat C_{\bm f,i}-C_{\bm f,i}\right\Vert^2_2\right]
\end{align}
which is the average of errors for each cell. \modified{Regarding the error of the coefficient, we define it between the value reconstructed from the predicted and ground-truth surface flow, so the error during training-data-generation can be eliminated. Then,} the error is actually a \textit{weighted} sum of \textit{signed} cell errors, where the weights can be roughly seen as the cell normals and areas according to equation \ref{eqn:force}. Since the pressure coefficients are the dominant part of the coefficient, we write out their contribution to the lift coefficient error: 

\begin{align}
    \mathcal L_{C_L}&= \mathrm{MSE} (\hat C_L; C_L)\nonumber\\
    &= \left(\sum_{i=1}^{N_{\mathrm{cell}}}(\hat C_{p,i}-C_{p,i})(\bm R_\alpha \cdot\bm n_i) \cdot A_i \right)^2 + \cdots
\end{align}
An unfavorable error distribution among cells or the offset of positive and negative errors will both lead a model with lower training loss in terms of surface flow, but due to the different scaling, will not necessarily result in an improved prediction in terms of aerodynamic coefficients. 

Hence, to raise the accuracy of the coefficients when the primary output is surface flow, we further add the prediction error of coefficients $\mathcal L_{\mathrm{coef}}= \mathrm{MSE} (\hat C_L; C_L)+\mathrm{MSE} (\hat C_D; C_D)$ to the loss function. The final loss function is

\begin{equation}\label{eqn:loss}
    \mathcal L = \mathcal L_{\mathrm{surf}} + \lambda \cdot \mathcal L_{\mathrm{coef}}
\end{equation}
where $\lambda$ is the weight of the aerodynamic loss term. In the following section, an experiment is conducted to decide the best $\lambda$. To investigate the effect of potentially conflicting update directions from the different loss terms, we also tested the hyperparameter-free ConFIG method~\cite{liu_config_2025} to combine the two loss terms.

\subsubsection{Model implementation}

\modified{For the hyperparameters of the AeroTransformer model we largely use those of the original PDE-Transformer.} \cite{holzschuh_pde-transformer_2025} \modified{Two down-sampling stages, one latent stage, and two corresponding up-sampling stages are used for the $\mathrm{AT}_\mathrm{surf}$ model. The number of blocks inside each stage is 2, 5, 8, 5, 2, contributing a total of 22 blocks. For $\mathrm{AT}_\mathrm{coef}$, the backbone only includes the down-sampling and the latent stages.

The selection of the patch sizes $p_H$ and $p_W$ for mesh patching and the window size $w$ for shifted-window attention is a trade-off between efficiency and model performance. Smaller $p_H$ and $p_W$ and a larger $w$ improve the model's representational ability, but at the cost of longer training time. Here, we use $p_H=p_W=4$ and $w=8$, as recommended by best practice in the PDE-Transformer paper} \cite{holzschuh_pde-transformer_2025}. \modified{Regarding the hyperparameters within the Transformer blocks, the initial hidden dimension $N_{\mathrm{hidden},0}$ controls the model size, and we compare three variants based on our training budgets in the following experiments, with $ N _ {\mathrm {hidden}, 0} $ set to 16, 32, and 64, corresponding to size S, M, and L. The number of heads and the MLP ratios are 8 and 4, following the common setting in Transformer-based models.}

\subsection{Training strategies}


\subsubsection{Pre-training} 

To enable stable, efficient training of the large model, an accurate, stable gradient is required. The AeroTransformer enables a larger mini-batch size thanks to its hierarchical architecture. In the following experiments, we test different batch sizes $B$. We also add gradient clipping \cite{pascanu_difficulty_2013} to avoid spikes in the learning curve. This is achieved by scaling the gradient vector's norm to 1 if it exceeds 1. We also attempted the exponential moving average (EMA) method for gradient clipping, as previously recommended~\cite {holzschuh_pde-transformer_2025}. However, this was detrimental to performance, as outlined in Appendix \ref{app:trainconfig}.\ref{app:clipping}. Additionally, the AdamW optimizer and a learning rate schedule based on the one-cycle policy are employed to update the parameters. Specifically, the learning rate increases from $0.04 \times LR_{\max}$ to $LR_{\max}$ in the first half of training and then decreases back to $0.001 \times LR_{\max}$ in the second half. The maximum learning rates ($LR_{\max}$) set to $1\times 10^{-3}$. 

\subsubsection{Fine-tuning} 

One of the key advantages of the two-stage training is that the model training cost for the second task-specific stage can be reduced, because the pre-trained model has already \modified{learned} the general trend and the deep physical features of the flow features in the design domain. Considering this, we tested different training settings to reduce the fine-tuning cost, i.e., smaller dataset, fewer training steps, and several parameter-efficient fine-tuning strategies below:

\paragraph{Fine-tuning only attention layers} Instead of updating all parameters in the pre-trained model, we tested updating only the query, key, and value projection matrices in the attention layers, while keeping all remaining parameters frozen. This is motivated by prior findings showing that they are the most influential for task-specific adaptation \cite{he_towards_2022}. We also tested including the input mesh-embedding layer and output projection layers in the update process, as the major differences between pre-training and fine-tuning stem from the geometric complexity of the shapes; however, this approach did not yield superior results.

\paragraph{Low rank adaptation (LoRA)} Beyond full-parameter tuning of these crucial layers, we further explore reducing the number of trainable parameters using LoRA \cite{hu_lora_2022}. For the existing weight matrices $\bm W \in \mathbb{R}^{d\times k}$ in the aforementioned crucial layers, LoRA injects a weight correction $\Delta \bm W$ that can be decomposed to a pair of lightweight low-rank matrices $\Delta \bm W=\bm W_1 \bm W_2$ where $\bm W_1 \in \mathbb{R}^{d\times r}$ and $\bm W_2 \in \mathbb{R}^{r\times k}$ are the only tunable parameters. The tuned weight matrices will be:

\begin{equation}
    \bm W' = \bm W + \frac{\alpha}{r} \bm W_1 \bm W_2
\end{equation}
where $\alpha$ is a scaling factor and we use $\alpha=2r$ as recommended. Since $r\ll d$, LoRA allows the model to approximate the fine-tuning direction with orders-of-magnitude fewer parameters while keeping the original weights frozen. Here, we follow the common setting and apply LoRA only to the \modified{query} and value projections \cite{hu_lora_2022}. The \modified{key} projection is frozen.

\paragraph{}

Another important consideration during fine-tuning is preserving the knowledge learned during pre-training. This, on one hand, can be achieved by the aforementioned freezing of certain parts of the parameters; on the other hand, we also lower the initial and maximum learning rate from $0.04 \times LR_{\max}$ and $10^{-3}$ to $0.005 \times LR_{\max}$ and $10^{-4}$, respectively. 

\section{Experiments on transonic wings}

Although the ultimate goal of our approach is to move toward a model that generalizes across a broad aerodynamic design space, the scope of the present study focuses on predicting aerodynamic coefficients and surface flow for three-dimensional transonic wings. For this task, we adopt the most diverse available transonic wing dataset and train a model that, to the best of our knowledge, is also the largest reported to date.

\modified{Through the following experiments, we demonstrate that our model exhibits several favorable properties typically associated with the foundation-model paradigm, particularly its effectiveness in supporting downstream tasks. Nevertheless, we emphasize that the current work does not claim to provide a universal model for arbitrary geometries or design spaces. Instead, the focus is on demonstrating the paradigm that how a model pre-trained on simplified shapes can be effectively adapted to more detailed downstream aerodynamic design tasks within a well-defined domain.}


\subsection{Problem definition}

\subsubsection{Shape parameters of transonic wings}

\begin{figure}[htbp]
    \centering
    \includegraphics[width=0.85\linewidth]{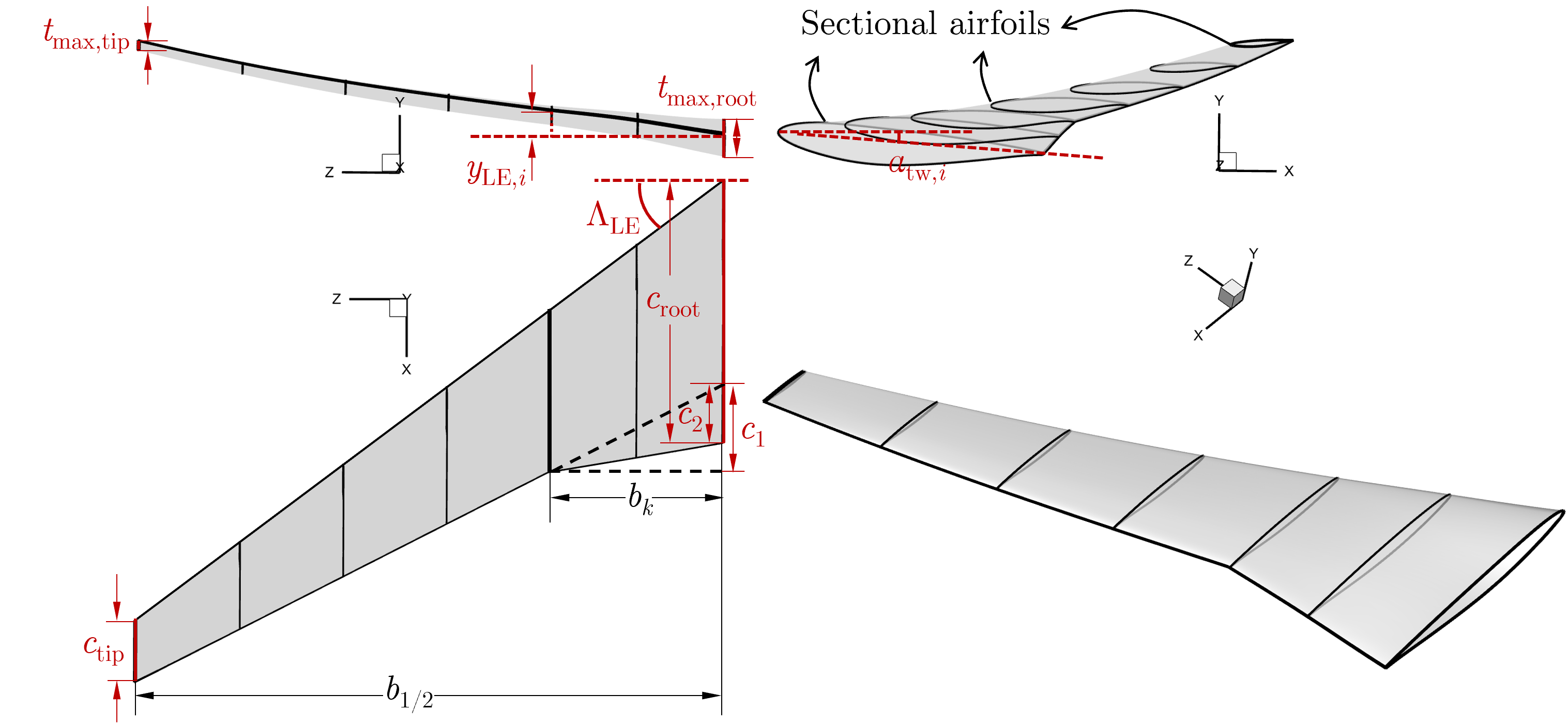}
    \caption{Transonic wing and its shape parameters}
    \label{fig:wingshape}
\end{figure}

Transonic wings are the crucial component of the modern transportation aircraft. A sketch of a typical wing is shown in Fig. \ref{fig:wingshape}. Despite its smooth surface, the wing shape actually has a large degree of freedom, most of which comes from the shapes of sectional airfoils. In modern transonic wings, the airfoil shapes at different spanwise stations can vary to achieve optimal performance. 

In addition to the airfoil shapes, there are \textit{global parameters} that control how the sectional airfoils form the wing. From the top view, a typical transonic wing follows a two-segment planform shape, and there is a break of the trailing edge, also known as the "kink". The parameters that determine the planform shape include the leading edge sweep angle $\Lambda_{\mathrm{LE}}$, aspect ratio $AR= 2b_{1/2}^2 / S_\mathrm{ref}$, taper ratio $TR=c_\mathrm{tip} / c_\mathrm{root}$, kink location $\eta_k=b_{1/2}/b_k$, and root adjustment ratio $\kappa=c_2/c_1$. In the formulas, we use the total projection area as the reference area; the definitions of the other parameters are shown in Fig. \ref{fig:wingshape}.

The planform shape determines the chord lengths and the locations of each sectional airfoil on the $x$-$z$ plane, while the spanwise distribution of $y$-axis locations and the rotation of the airfoils along $z$-axis are further determined by two extra group of parameters, the dihedrals $y_{\mathrm{LE}}$ and the twist angles $\alpha_{\mathrm{tw}}$. 

\subsubsection{Data formats} 

The wing flow field prediction task in this paper is to predict wing surface flow fields for various wing shapes and operating conditions. 

\begin{figure}[ht]
    \centering
    \includegraphics[width=0.6\linewidth]{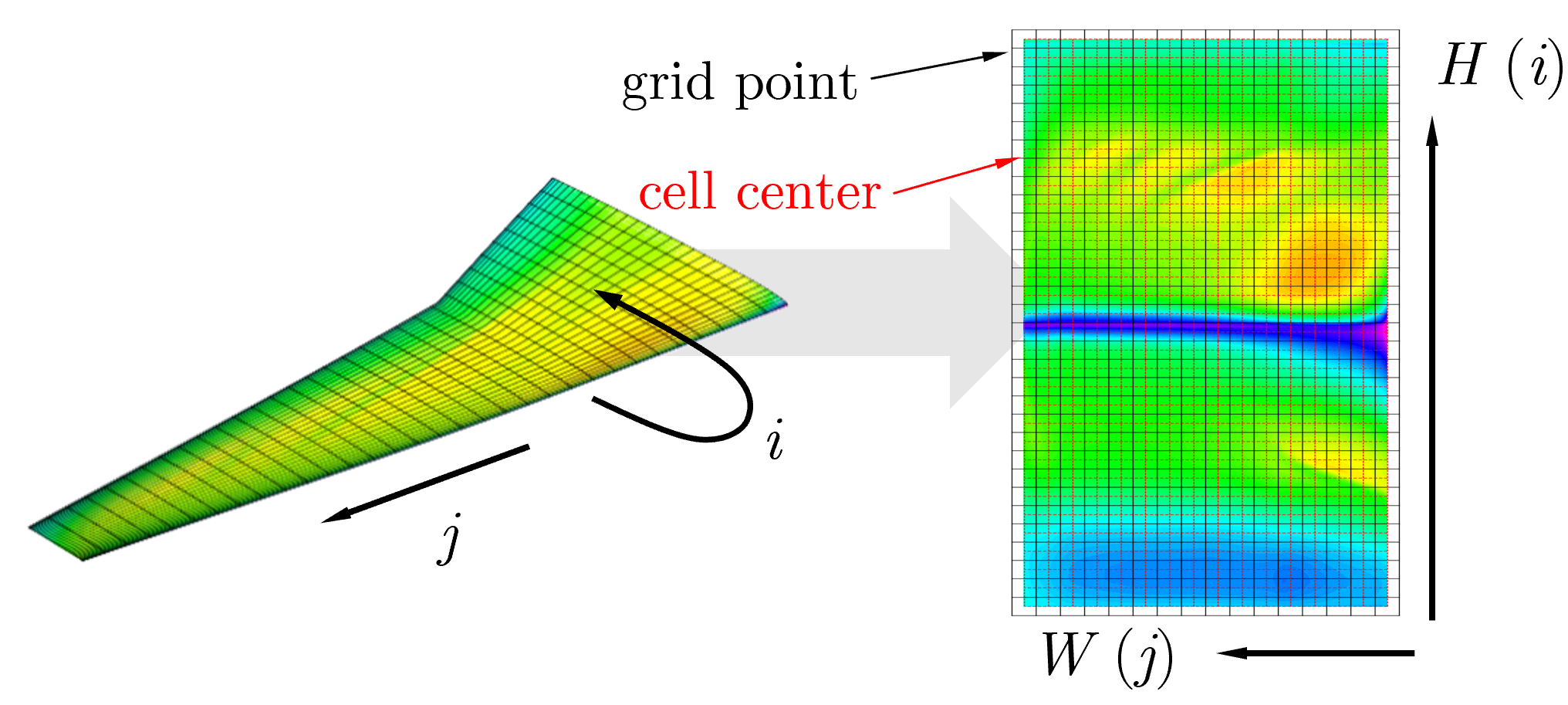}
    \caption{Transferring surface mesh and quantities from simulation mesh to reference mesh}
    \label{fig:meshshape}
\end{figure}

\paragraph{Input: Surface mesh} AeroTransformer takes a structured mesh as input, and a wing's surface can be naturally described with a structured mesh. As illustrated in Fig. \ref{fig:meshshape}, the mesh has one dimension circled around the wing and the other on the spanwise. In this paper, we use the same surface-mesh prototype for all samples. In the spanwise ($j$-direction), \modified{a very small region near the wing tip is removed during post-processing to construct a consistent structured mesh}\footnote{\modified{The removed region is $\approx 0.3\%$ of the span. When reconstructing aerodynamic coefficients from structured-mesh data, this part's contribution is recovered by interpolating the quantities to the corresponding streamwise positions at the upper and lower surfaces. This treatment, as well as the interpolation, only introduce negligible difference ($\approx 0.1\%$) in the integrated aerodynamic coefficients. }}, and the rest of the part is discretized with 129 evenly spaced cross-sections. For each cross-section ($i$-direction), a fixed set of normalized chordwise positions is used for both the upper and lower surfaces, and the tail edge is represented only with one cell. The model input uses the cell-centered coordinates of its mesh, which are matrices of size $256 \times 128 \times 3$.  

\paragraph{Input: Operating condition} The input of operating conditions involves two quantities, i.e., the Mach number $Ma$ and the angle of attack $\alpha$. Other freestream quantities, including the Reynolds number and temperature, are set to 20 million and \modified{300 $^\circ$ K}, respectively, as they have a lesser effect on the resulting flow field. 

\paragraph{Outputs: Surface flow field and aerodynamic coefficients} As mentioned in the last section, the model outputs are two-fold. The surface flow quantities, including the pressure coefficient $C_p$ and the friction coefficient $C_{\bm f}$, are on the grid center coordinates that are the same as the geometric input. Since $C_{\bm f}$ is a vector tangential to the surface, it is decomposed into streamwise and spanwise components. The aerodynamic coefficients include the lift coefficient $C_L$, the drag coefficient $C_D$, and the pitching momentum coefficient $C_{M,z}$.

\subsubsection{Pre-training and fine-tuning datasets}

We employ two datasets with different purposes to demonstrate both the scalability of the AeroTransformer model and the effectiveness of the two-stage methodology. The model is first pre-trained on the existing SuperWing dataset \cite{superwing}, which provides large-scale geometric diversity. A new task-specific fine-tuning dataset is constructed in this work to evaluate the model’s ability to adapt to a specific shape-optimization problem. \modified{A quantitative summary of their parameter ranges and sampling strategies is provided in Table}~\ref{tab:datasets}.

\paragraph{Pre-training dataset}

As outlined above, the majority of a wing’s geometric degrees of freedom arise from its sectional airfoil shapes. To enable efficient sampling of a broad design space while maintaining representative aerodynamic behavior, the SuperWing dataset uniformly samples all global parameters ($\Lambda_{\mathrm{LE}}$, $AR$, $TR$, $\eta_k$, and $\kappa$) while simplifies the spanwise distributed parameters (sectional airfoil shape, $y_{\mathrm{LE}}$, and $\alpha_{\mathrm{tw}}$) with certain \textit{patterns}. 

Sectional airfoil shapes are generated by modifying a single baseline airfoil, with variations introduced through changes in maximum thickness and camber along the spanwise direction. The spanwise distributions of maximum thickness, camber, dihedral, and twist are represented using B-spline curves defined by five control points. This parameterization significantly reduces the design space's dimensionality while preserving the dominant aerodynamic trends.

\modified{For each wing, eight operating conditions are randomly sampled from the range of $Ma\in [0.75, 0.90]$ and $AoA \in [2^\circ, 12^\circ]$, given there is no installation angle. The Reynolds number and freestream temperature are fixed at 20 million and 300 $^\circ$ K, respectively. By collecting the convergence results, the Superwing dataset comprises 4239 distinct wing shapes and 28856 flow fields. Further details of the dataset construction can be found in Ref}.~\cite{superwing}. 

\paragraph{Fine-tuning dataset}

The fine-tuning dataset is designed to train a relatively local surrogate model tailored to a specific aerodynamic shape-optimization task. Unlike the pre-training dataset, this dataset focuses not on global geometric diversity but on capturing more detailed variations with richer shape parameters within the design neighborhood. 

In this work, we construct a task-specific dataset based on the NASA Common Research Model (CRM) wing \cite{vassberg_development_2008}, which serves as the baseline configuration. Reflecting common industrial optimization setups, the wing geometry is controlled at seven spanwise sections (located as shown in Fig. \ref{fig:wingshape}), where sectional airfoils are allowed to vary independently rather than being derived from a single baseline shape. At each control section, the sectional airfoil is parameterized using 20 CST coefficients, each perturbed within $\pm 40\%$ of the corresponding CRM coefficient. Dihedral offsets at the control sections are perturbed within $\pm 0.05 \cdot c_{\mathrm{root}}$, and twist angles are sampled uniformly from $-3^\circ$ to $0^\circ$. The spanwise distributions of these parameters are then constructed via interpolation between the control sections. Compared to the pre-training dataset, the fine-tuning dataset has greater degrees of freedom (153 versus 37), which aligns with the common design task. 

\modified{For each perturbed wing, we also select eight operating conditions. The $Ma$ range remains the same with SuperWing, while we narrow the $AoA$ to $[-2^\circ, 4^\circ]$, given the installation angle is $6.71^\circ$. This procedure yields a fine-tuning dataset comprising 288 wing geometries and 2145 corresponding flow fields, but we will demonstrate in the following sections that using only a small subset of the data can yield good task-specific results}

\paragraph{}

\begin{table}[htbp]
    \centering
    \caption{Details of the pre-training and fine-tuning dataset}
    \label{tab:datasets} 
    \begin{tabular}{cccM{4cm}M{4cm}}
        \hline \hline
        \multicolumn{3}{c}{Dataset} & \makecell{Pre-training\\(SuperWing)} & \makecell{Fine-tuning\\(Task-specific, CRM)} \\
        \midrule
        \multirow{2}{*}{Amount of} & \multicolumn{2}{c}{Shapes} & 4239 & 288\\
        & \multicolumn{2}{c}{Flow fields} & 28856 & 2145\\
        \multirow{8}{*}{\makecell{Sampling \\ strategy}} & \multirow{5}{*}{\makecell{Wing-\\specific \\ parameters }} & $\Lambda_\mathrm{LE}$ & $\mathcal U [25^\circ, 40^\circ]$ & fixed $37.16^\circ$ \\ 
        && $AR$ & $\mathcal U [8, 11]$ & fixed 8.38\\
        && $TR$ & $\mathcal U [0.15, 0.40]$ & fixed 0.275\\
        && $\eta_k$ & $\mathcal U [36\%, 42\%]$ & fixed 36.8\%\\
        && $\kappa_\mathrm{root}$ & $\mathcal U [10\%, 110\%]$ & fixed 67.0\% \\
        & \multirow{3}{*}{\makecell{Spanwise-\\ distributed \\ parameters }} & \makecell{Sectional \\ airfoil \\ shapes} & \makecell[l]{\parbox[t]{4cm}{one baseline airfoil described with 20 CST coefficients; maximum thickness and camber vary along spanwise according to a spline built by 5 control points}} & \makecell[l]{\parbox[t]{4cm}{linearly interpolated from 7 control sectional airfoils described each with 20 CST coefficients; each coefficient perturbed from CRM}}\\
        && $y_{\mathrm{LE}}$s & \makecell[l]{\parbox[t]{4cm}{spline built by 2 points at kink and tip}} & \multirow{2}{*}{\makecell[l]{\parbox[t]{4cm}{spline built by 7 control points, where their values perturbed from CRM}}}\\
        && $\alpha_{\mathrm{tw}}$s & \makecell[l]{\parbox[t]{4cm}{spline built by 5 control points}}  & \\
        \multicolumn{3}{c}{Degrees of freedom} & 37 & 153 \\
        \hline \hline
    \end{tabular}
\end{table}

Figure \ref{fig:crmdisp} visually demonstrates the shapes in the two datasets and highlights their differences. \modified{On the top left, the wing planform shapes are displayed from the top view, where 5\% wings from the pre-training SuperWing dataset are plotted and cover a broad range. The fine-tuning CRM dataset, on the other hand, only has the same platform shape as marked red. On the right, a $Ma$-$AoA$ plot shows the operating conditions for the two datasets. We note that for higher Mach numbers, strong shock waves may lead to reduced convergence robustness in CFD simulations, resulting in a relatively sparser distribution of valid samples in this region. 

In the bottom subfigure, sectional airfoil shapes are compared. For each wing, normalized sectional airfoil shapes (solid lines) and camber lines (dashed lines) are shown at six spanwise stations. We provide a stack of airfoils over the entire dataset at the first slot, and we also show the airfoils from five randomly-selected wings.} The sectional airfoils also exhibit markedly larger variations, particularly near the root region, where the CRM has a negative camber. As a result, the fine-tuning dataset extends beyond the parametric envelope of the pre-training data, making the transfer from the pre-trained model to this task challenging.

\modified{In Appendix} \ref{app:dim} \modified{, we further present dimension-reduction results for the two datasets to better understand the relationship between their intrinsic dimensionality.}

\begin{modifiedblock}
\begin{figure}[H]
    \centering
    \begin{minipage}{0.4\textwidth}
        \centering
        \begin{subfigure}{\linewidth}
            \centering
            \includegraphics[width=\linewidth]{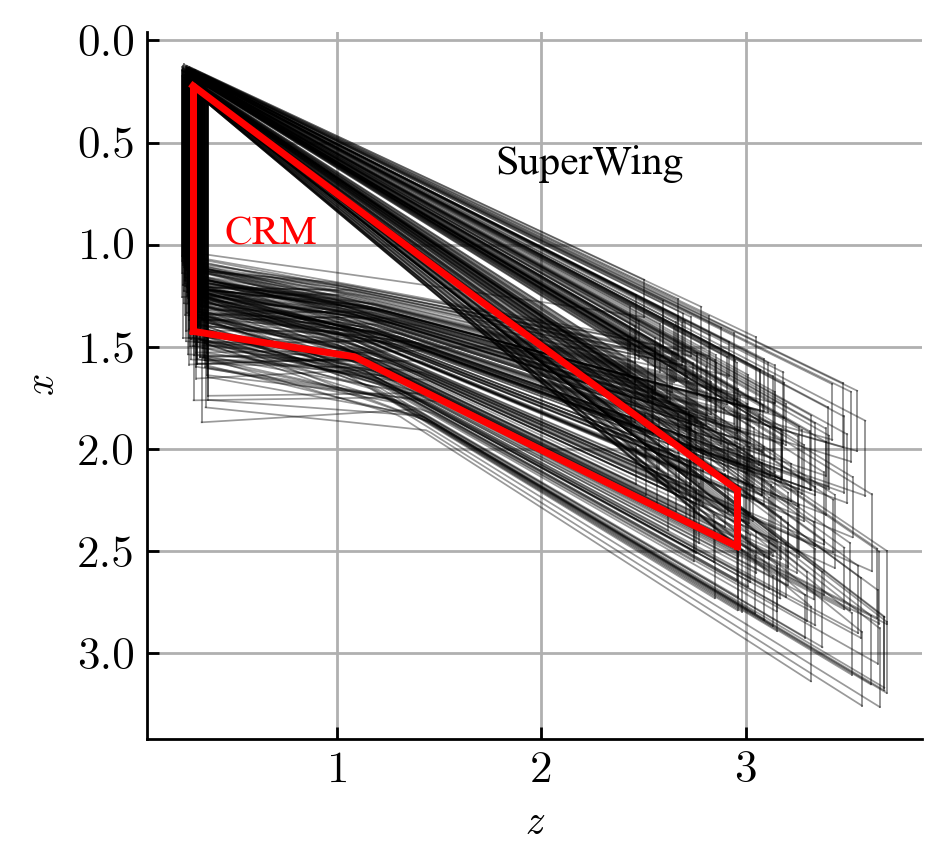}
            \caption{Planform shapes}
            \label{fig:planform}
        \end{subfigure}
    \end{minipage}
    \hspace{0.5em}
    \begin{minipage}{0.4\textwidth}
        \centering
        \begin{subfigure}{\linewidth}
            \centering
            \includegraphics[width=\linewidth]{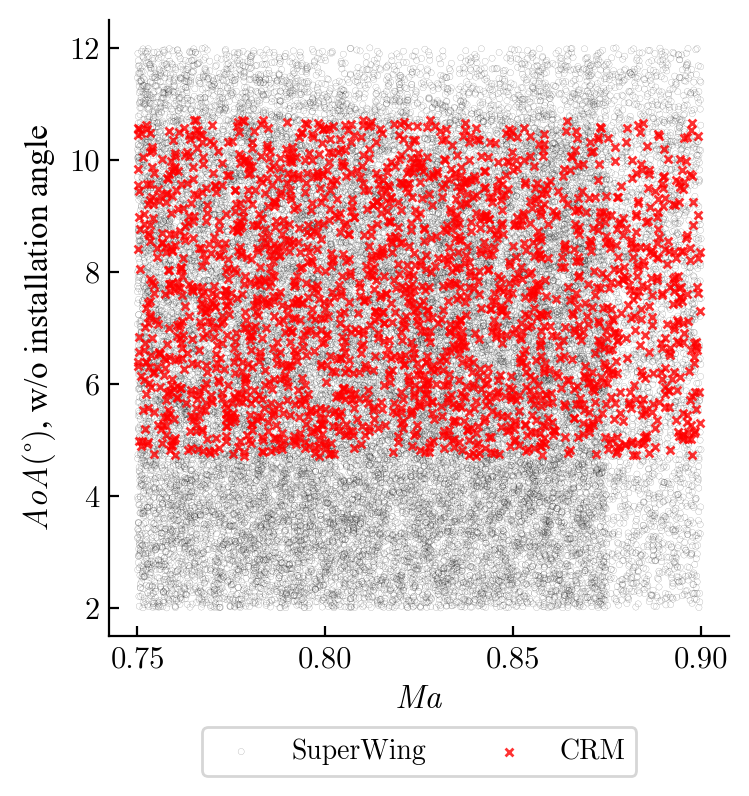}
            \caption{Operating conditions}
            \label{fig:ocs}
        \end{subfigure}
    \end{minipage}
    \vspace{0.5em}

    \begin{subfigure}{\linewidth}
        \centering
        \includegraphics[width=0.85\linewidth]{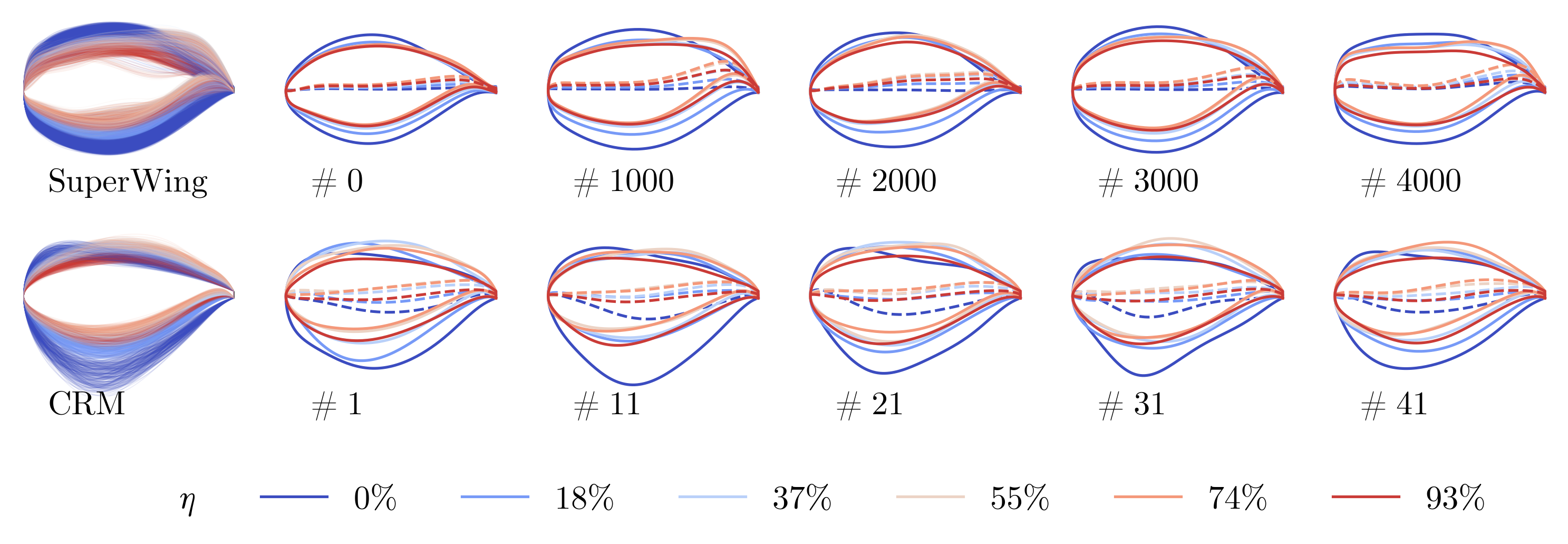}
        \caption{Sectional airfoils}
        \label{fig:sections}
    \end{subfigure}

    \caption{An illustration highlighting the substantial differences in pre-training dataset (SuperWing) and fine-tuning dataset (CRM)}\label{fig:crmdisp}
\end{figure}
\end{modifiedblock}

\subsubsection{Solver}

\modified{Reynolds-Averaged Navier-Stokes (RANS) simulations are performed using the open-source CFD solver suite developed by the} \verb|MDOLab| \footnote{https://mdolab.engin.umich.edu/software}. \modified{The surface mesh generation is with in-house CAD code and the volume mesh using} \verb|pyHyp|. \modified{The flow fields are computed using} \verb|ADflow| \cite{mader_adflow_2020}. \modified{A “3w” multigrid strategy is adopted to accelerate convergence, with a total of 4,000 iterations. Convergence is declared when the residual falls below $1\times10^{-10}$. All other solver settings follow the default configuration. In our previous paper on the dataset} \cite{superwing}, \modified{we provided a mesh convergence study to validate the simulation.} The simulations are conducted on a 160-processor cluster of Intel Xeon Gold 5320 CPUs.

\subsubsection{Error measurements for the surrogate model}

For each sample (a wing shape under one operating condition), the relative mean absolute error (MAE) is computed for the surface pressure and friction coefficients and normalized by the corresponding coefficient range. The final test error is reported as the average of these normalized MAEs \modified{in percentage} across all test samples, as follows:

\begin{modifiedblock}
\begin{equation}
    \delta X = \frac{1}{N_s} \sum_{n=1}^{N_s}\frac{\frac{1}{H\cdot W}\sum_{i,j=1}^{H,W}\left|\hat X_{n;i,j}-X_{n;i,j}^\mathrm{CFD}\right|}{\max_{i,j}X_{n;i,j}^\mathrm{CFD}-\min_{i,j}X_{n;i,j}^\mathrm{CFD}} \times 100\%, \quad X\in \left[C_p, C_{f,\tau}, C_{f,z}\right]
\end{equation}
\end{modifiedblock}
\noindent where $N_s$ is the number of samples. \modified{To enable a more concise visualization in the following plots, we introduce an aggregated metric, \textit{surface flow error}, which is calculated by averaging the normalized MAEs in the three channels:}

\begin{modifiedblock}
\begin{equation}
    SFE = \frac{1}{3}\left(\delta C_p + \delta C_{f, \tau} + \delta C_{f,z}\right) \times 100\%
\end{equation}
\end{modifiedblock}

In addition to surface field errors, the mean absolute errors in key aerodynamic coefficients, including lift, drag, and pitching moment about the leading edge, are also evaluated to further validate predictive accuracy. 

\begin{modifiedblock}
\begin{equation}
    \delta X = \frac{1}{N_s} \sum_{n=1}^{N_s}\left|\hat X_n-X_n^\mathrm{CFD}\right|, \quad X\in \left[C_L, C_D, C_{M,z}\right]
\end{equation}
\end{modifiedblock}

\subsection{Model pre-training}

In this section, we test the AeroTransformer on the aforementioned pre-training dataset. 

\subsubsection{Overall performance}

\paragraph{Architectural improvement}

\modified{We compare the surface flow version of S-size AeroTransformer with the baselines. It has two variants: one follows the common practice of inputting the operating conditions together with the mesh, and the other follows the operating condition injection method proposed in this paper.} The baselines include the commonly used U-Net \cite{yang_transferable_2024}, the original Vision Transformer \cite{dosovitskiy_vit_2022}, and the Transolver \cite{wu_transolver_2024}. \modified{We select the hyperparameter of the baseline models based on our previous test on a smaller dataset} \cite{yang_rapid_2025, yang_eucass_2025} \modified{and ensure they have a similar trainable parameter count of 1.0M. Details of their implementation are provided in Appendix} \ref{app:baseline}.  

The training is conducted on an NVIDIA RTX A5000. \modified{Given memory constraints, a training batch size of 4 was used for all models.} 
The models are trained on 90\% of the pre-training dataset, while the remaining 10\% is used for testing. To ensure robustness, the training process is repeated three times, using different random subsets of 90\% of the training data to cross-validate the results. The error scores in Table \ref{tab:modperf} represent the averages across the three cross-validation runs. 

\begin{table}[ht]
\small
    \centering
    \caption{\label{tab:modperf}Model performance comparison}
    \begin{tabular}{cccccccccc}
        \hline\hline
        \multirow{2}{*}{Model} & \multicolumn{6}{c}{Errors} & \multirow{3}{*}{\makecell{\# of\\ Param.}} & \multirow{2}{*}{\makecell{Training \\ time}} & \multirow{2}{*}{\makecell{Peak \\ memory}} \\ 
        & $\delta C_p$ & $\delta C_{f,\tau}$ & $\delta C_{f,z}$ & $\delta C_L$ & $\delta C_D$ & $\delta C_{M,z}$\\
        & \multicolumn{3}{c}{\mdf Percentage relative error (\%)} & \mdf ($\times 10^{-3}$) & \mdf ($\times 10^{-4}$) & \mdf ($\times 10^{-3}$) & & (hours) & (GB)\\
        \midrule
        U-Net & 0.911 & 0.548 & 0.635 &  17.85 & 11.71 & 19.28 & 2.3M & 15.3 & 1.03\\ 
        ViT & \underline{0.353} & \underline{0.267} & \underline{0.313} &  \underline{2.90} & \underline{2.60} & \underline{3.26} & 1.2M & 20.0 & 4.61 \\ 
        Transolver & 0.401 & 0.309 & 0.357 &  3.15 & 3.00 & 3.62 & 1.0M & 44.6 & 10.71 \\ 
        AeroTransformer  & 0.279 & 0.209 & 0.253 &  2.44 & 2.21 & 2.86 & 1.0M & 18.8 & 0.62\\
        \makecell[l]{ $\quad + ~ $ o.c. injection}  & \textbf{0.264} & \textbf{0.198} & \textbf{0.238} &  \textbf{2.35} & \textbf{2.11} & \textbf{2.75} & 1.0M & 18.8 & 0.64\\
        \hline\hline 
    \end{tabular}
\end{table}

Compared to ViT, AeroTransformer introduces a hierarchical architecture with local attention, enabling deeper models while reducing parameter count and memory consumption. This design leads to consistent improvements across all error metrics, achieving approximately a 15\% relative improvement over ViT. Further gains are observed when operating condition injection is included, yielding an additional 4\% improvement. 

\begin{figure}[ht]
    \centering
    \includegraphics[width=0.8\linewidth]{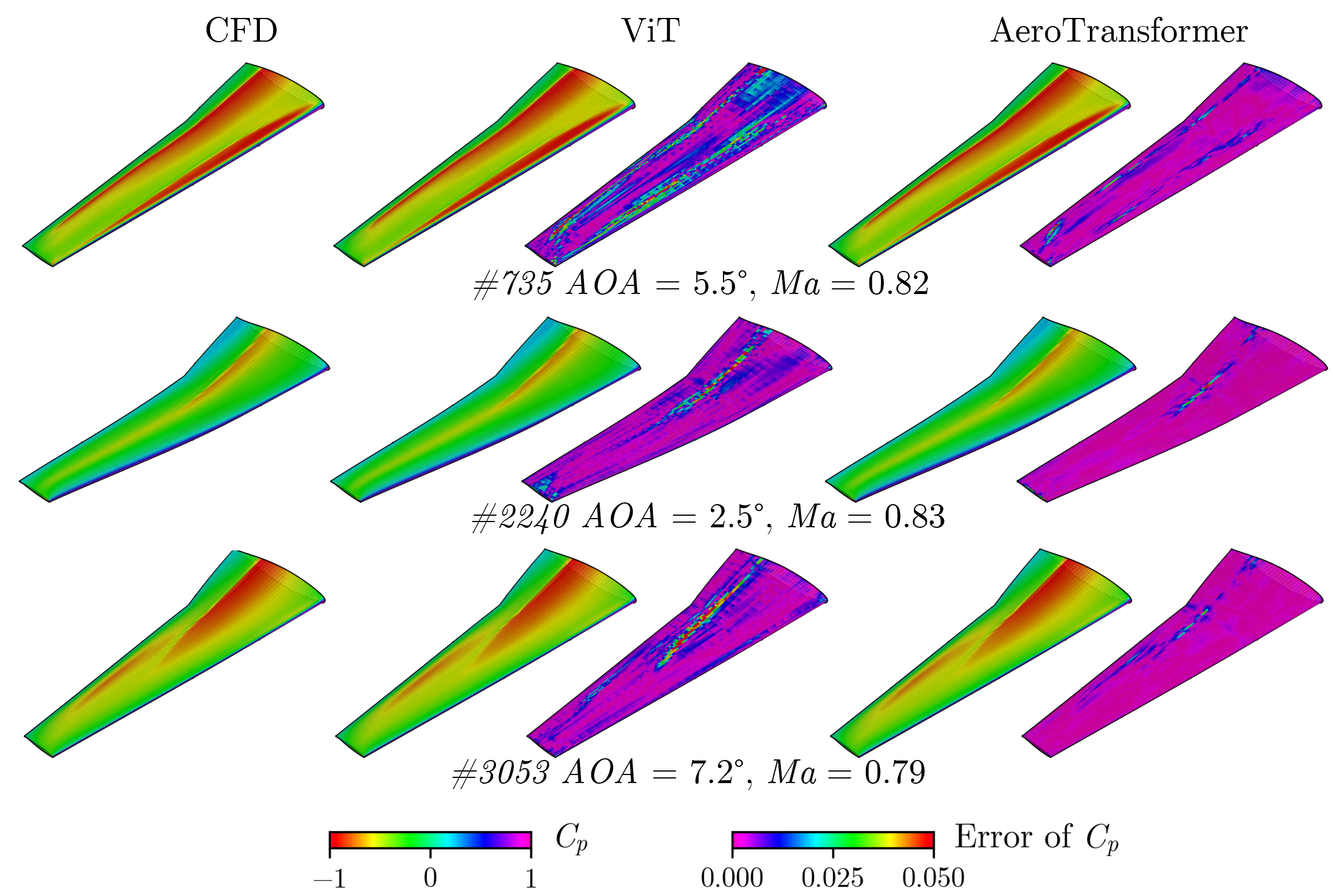}    \caption{Surface pressure predictions from ViT and AeroTransformer}
    \label{fig:pretrainfield}
\end{figure}

In Fig. \ref{fig:pretrainfield}, we provide an intuitive comparison of the pressure coefficient fields predicted by ViT and AeroTransformer on three samples selected from the unseen testing dataset. Although both models produce good pressure contours, the error distribution indicates that AeroTransformer has better accuracy in capturing the field's details. The errors near regions of strong variation (such as the shock wave) are reduced.

\paragraph{Trained with larger batch sizes}

\modified{Besides the primary improvement in the model architecture, AeroTransformer's reduced memory usage enables further scaling. On the one hand, this allows for larger model sizes and higher input resolutions; on the other hand, it permits the use of larger batch sizes under the same hardware constraints, which can stabilize the optimization process and improve final performance.

In Fig.} \ref{fig:scalebatch}, \modified{we report the performance of the S-size AeroTransformer trained with increasing batch size $B$, while keeping the total number of optimization steps fixed.\footnote{The experiments with the largest batch sizes, 256, were conducted on an NVIDIA L40 GPU and, due to their high computational cost, were performed with a single run.}. The results exhibit a decrease in prediction errors up to $B=64$, with an approximately linear trend with respect to the logarithm of the batch size. Beyond this point, the performance gain becomes marginal. We also report the total training time per run and peak memory usage, both of which increase approximately linearly with batch size. In all, larger batches primarily offer benefits in final performance but come at the cost of longer training time.} 

\begin{modifiedblock}
\begin{figure}[H]
\centering
\small
\begin{minipage}{0.35\textwidth}
    \centering
    \begin{subfigure}{0.9\linewidth}
    \begin{tikzpicture}
    \begin{axis}[
      width=\textwidth,
      height=0.7\textwidth,
      axis lines*=left,
      xmin=3, xmax=260,
      xmode=log,
      xtick={4, 16, 64, 256},
      xticklabels={$4$, $~~~16$, $~64$, $256^*$},
      xlabel={Batch sizes $B$},
      ylabel={Surface flow error ($SFE$)},
      yticklabel={
            \pgfmathparse{\tick}
            \pgfmathprintnumber[fixed,fixed zerofill,precision=2]{\pgfmathresult}\%
        },
    ]
    
    \addplot+[color=black, mark=o, mark size=3pt,
      error bars/.cd, y dir=both,y explicit,]
    coordinates {
    (4, 0.233333) +- (0, 0.00295)
    (16, 0.17578) +- (0, 0.00442)
    (64, 0.14149) +- (0, 0.00365)
    (256, 0.13898)
    };

    \end{axis}
    \end{tikzpicture}
    \caption{Prediction errors on the testing dataset}
    \end{subfigure}
\end{minipage}
\hspace{0.5em}
\begin{minipage}{0.3\textwidth}
    \centering
    \begin{subfigure}{0.9\linewidth}
    \begin{tikzpicture}
    \begin{axis}[
      width=\textwidth,
      height=0.8\textwidth,
      axis lines*=left,
      xmin=3, xmax=260,
      xtick={4, 16, 64, 256},
      xticklabels={$4$, $~~~16$, $~64$, $256^*$},
      xlabel={Batch sizes $B$},
      ylabel={Training time (hours)},
      ylabel style={yshift=-15pt},
    ]
    
    \addplot+[color=black, mark=o, mark size=3pt]
    coordinates {
    (4, 18.806)
    (16, 31.8158)
    (64, 48.9)
    (256, 127.1)
    };

    \end{axis}
    \end{tikzpicture}
    \caption{Training times with full pre-training dataset}
    \end{subfigure}
\end{minipage}
\hspace{0.5em}
\begin{minipage}{0.3\textwidth}
    \centering
    \begin{subfigure}{0.9\linewidth}
    \begin{tikzpicture}
    \begin{axis}[
      width=\textwidth,
      height=0.8\textwidth,
      axis lines*=left,
      xmin=3, xmax=260,
      xtick={4, 16, 64, 256},
      xticklabels={$4$, $~~~16$, $~64$, $256^*$},
      xlabel={Batch sizes $B$},
      ylabel={Peak memory (GB)},
      ylabel style={yshift=-15pt},
    ]
    
    \addplot+[color=black, mark=o, mark size=3pt]
    coordinates {
    (4, 0.617)
    (16, 2.326)
    (64, 9.047)
    (256, 35.9)
    };

    \end{axis}
    \end{tikzpicture}
    \caption{Training peak memory}
    \end{subfigure}
    \end{minipage}
\caption{AeroTransformer performance when trained with larger batch sizes}\label{fig:scalebatch}
\end{figure}
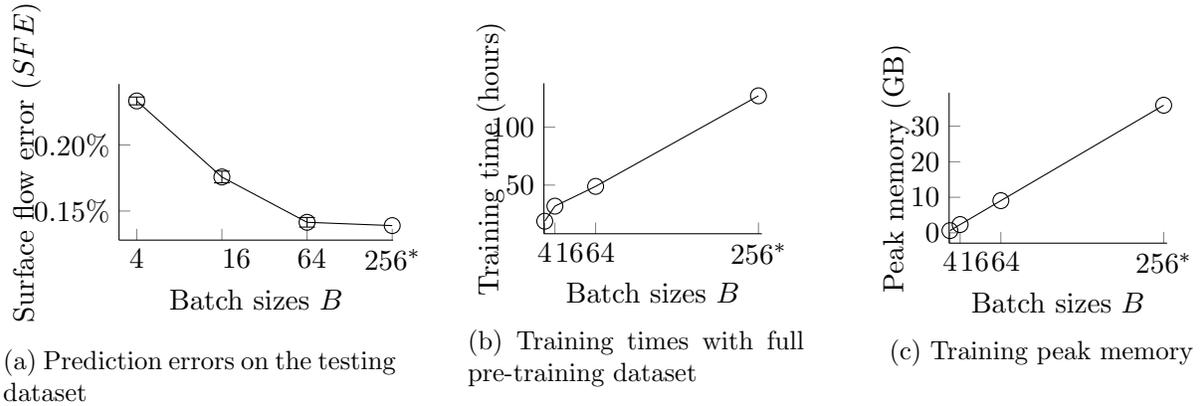
\end{modifiedblock}

\subsubsection{Scalability}\label{sec:scale}

To examine the scalability of the proposed model, we further investigate how performance changes as both the model capacity and the amount of pre-training data increase. In addition to the S-size configuration, we construct two larger variants of the AeroTransformer, denoted as M and L, whose initial token dimensions in the first stage are 32 and 64, corresponding to approximately 3.8 M and 10.5 M trainable parameters, respectively. For each model size, we train on 50\%, 75\%, and 90\% of the pre-training dataset, while maintaining the held-out test split identical to that used in the previous experiments. 

All models are trained for 585.6k optimization steps, \modified{which means} runs using only a fraction of the pre-training data effectively see more epochs per available sample. The batch size $B$ is set to 64 for all training runs. To stabilize optimization of the largest model, which exhibits loss oscillations under the default settings, we reduce the gradient clipping norm from 1.0 to 0.5, following common practice, and we also find that a smaller norm does not help (see Appendix \ref{app:trainconfig}.\ref{app:clipping}). 

We report the mean and standard deviation of the final prediction errors in Fig. \ref{fig:scaleall} (a). \modified{The corresponding training wall time and peak memory usage for different model sizes are shown in Fig.} \ref{fig:scaleall} \modified{(b) and (c), respectively. Plus, the inference time reported in Fig.} \ref{fig:scaleall} \modified{(b) is measured using random inputs with $B = 9$, which corresponds to evaluating the aerodynamic performance of a single geometry under nine operating conditions.}

\begin{modifiedblock}
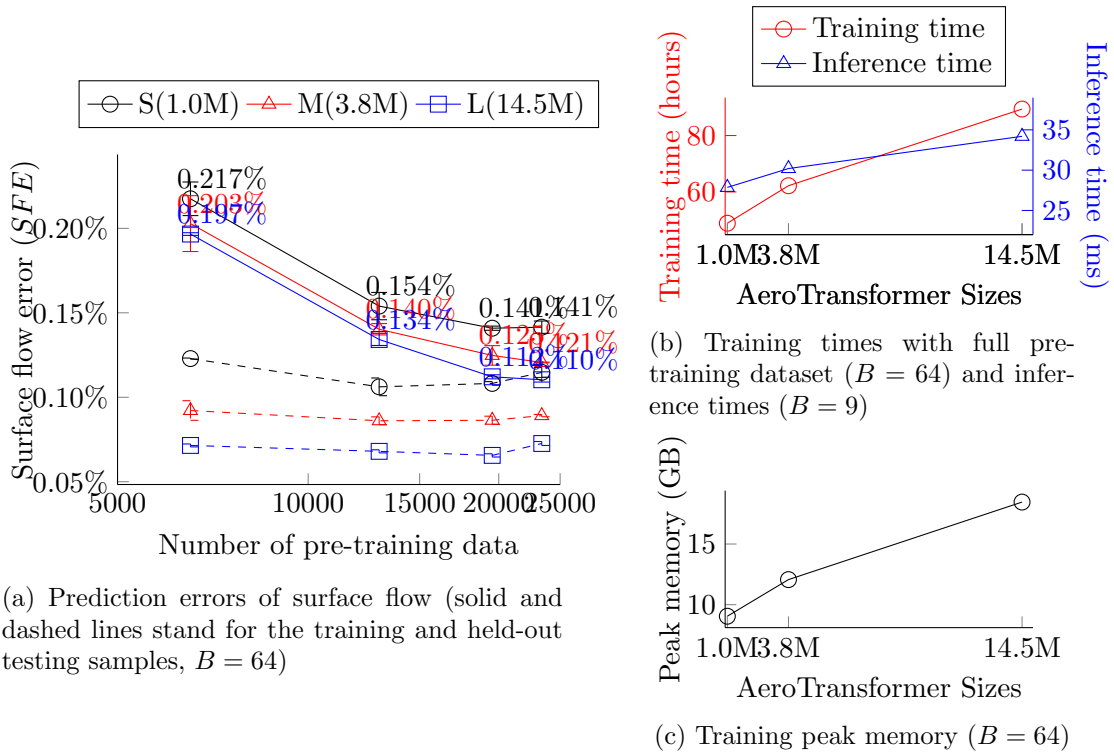
\begin{figure}[H]
\centering
\small
\begin{minipage}{0.5\textwidth}
    \centering
    \begin{subfigure}{0.9\linewidth}
    \begin{tikzpicture}
    \begin{axis}[
      width=\textwidth,
      height=0.8\textwidth,
      axis lines*=left,
      xmin=5000, xmax=25000,
      xmode=log, 
      xtick={5000, 10000, 15000, 20000, 25000},
      xticklabels={5000, 10000, 15000, 20000, 25000},
      xlabel={Number of pre-training data},
      ylabel={Surface flow error ($SFE$)},
      yticklabel={
            \pgfmathparse{\tick}
            \pgfmathprintnumber[fixed,fixed zerofill,precision=2]{\pgfmathresult}\%
        },
      legend columns=3,
      legend style={
        at={(0.5,1.08)},
        anchor=south,
        cells={anchor=west},
      },
    ]
    
    \addplot+[
      color=black,
      mark=o,
      mark size=3pt,
      nodes near coords={
        \pgfmathprintnumber[fixed,fixed zerofill,precision=3]{\pgfplotspointmeta}\%
      },
      nodes near coords style={
        xshift=12pt,
        },
      error bars/.cd, y dir=both,y explicit,
    ]
    coordinates {
    (6519, 0.21749) +- (0, 0.00995)
    (12946, 0.15397) +- (0, 0.00809)
    (19535, 0.14089) +- (0, 0.00124)
    (23386, 0.14149) +- (0, 0.00365)
    };
    
    \addplot+[
      color=black,
      dashed,
      mark=o,
      mark size=3pt,
      mark options={solid, fill=none},
      error bars/.cd, y dir=both,y explicit
    ]
    coordinates {
    (6519, 0.12290) +- (0, 0.00031)
    (12946, 0.10617) +- (0, 0.00525)
    (19535, 0.10820) +- (0, 0.00105)
    (23386, 0.11474) +- (0, 0.00288)
    };
    
    \addplot+[
      color=red,
      mark=triangle,
      mark size=3pt,
      nodes near coords={
        \pgfmathprintnumber[fixed,fixed zerofill,precision=3]{\pgfplotspointmeta}\%
      },
      nodes near coords style={
        xshift=12pt,
        },
      error bars/.cd, y dir=both,y explicit,
    ]
    coordinates {
    (6519, 0.20281) +- (0, 0.01654)
    (12946, 0.14025) +- (0, 0.00336)
    (19535, 0.12478) +- (0, 0.00579)
    (23386, 0.12059) +- (0, 0.00023)
    };
    
    \addplot+[
      color=red,
      dashed,
      mark=triangle,
      mark size=3pt,
      mark options={solid, fill=none},
      error bars/.cd, y dir=both,y explicit,
    ]
    coordinates {
    (6519, 0.09216) +- (0, 0.00580)
    (12946, 0.08608) +- (0, 0.00232)
    (19535, 0.08638) +- (0, 0.00242)
    (23386, 0.08913) +- (0, 0.00068)
    };
    
    \addplot+[
      color=blue,
      mark=square,
      mark size=3pt,
      nodes near coords={
        \pgfmathprintnumber[fixed,fixed zerofill,precision=3]{\pgfplotspointmeta}\%
      },
      nodes near coords style={
        xshift=12pt,
        },
      error bars/.cd, y dir=both,y explicit,
    ]
    coordinates {
    (6519, 0.19662) +- (0, 0.00079)
    (12946, 0.13420) +- (0, 0.00353)
    (19535, 0.11211) +- (0, 0.00120)
    (23386, 0.11036) +- (0, 0.00136)
    };
    
    \addplot+[
      color=blue,
      dashed,
      mark=square,
      mark size=3pt,
      mark options={solid, fill=none},
      error bars/.cd, y dir=both,y explicit,
    ]
    coordinates {
    (6519, 0.07148) +- (0, 0.00087)
    (12946, 0.06798) +- (0, 0.00092)
    (19535, 0.06559) +- (0, 0.00096)
    (23386, 0.07263) +- (0, 0.00118)
    };
    
    \legend{S(1.0M), , M(3.8M), , L(14.5M), }
    \end{axis}
    \end{tikzpicture}
    \caption{Prediction errors of surface flow (solid and dashed lines stand for the training and held-out testing samples, $B = 64$)}
    \end{subfigure}
    \end{minipage}
    \hspace{0.5em}
    \begin{minipage}{0.38\textwidth}
    \centering
    \begin{subfigure}{0.9\linewidth}
    \begin{tikzpicture}
    \begin{axis}[
      width=\textwidth,
      height=0.6\textwidth,
      axis lines*=left,
      xmin=0.9, xmax=15,
      xtick={1, 3.8, 14.5},
      xticklabels={1.0M, 3.8M, 14.5M},
      xlabel={AeroTransformer Sizes},
      ylabel={Training time (hours)},
      ylabel style={red, yshift=-15pt},
      y axis line style={red},
      y tick style={red},
      yticklabel style={text=red},
    ]
    
    \addplot+[color=red, mark=o, mark size=3pt]
    coordinates {
    (1.0, 48.9)
    (3.8, 62.2)
    (14.5, 89.4)
    };

    \end{axis}
    
    \begin{axis}[
      width=\textwidth,
      height=0.6\textwidth,
      axis x line*=none,
      axis y line*=right,
      xmin=0.9, xmax=15,
      ymin=22,ymax=39,
      xtick={1, 3.8, 14.5},
      xticklabels={1.0M, 3.8M, 14.5M},
      xlabel={AeroTransformer Sizes},
      ylabel={Inference time (ms)},
      ylabel style={blue,
        at={(axis description cs:1.5,0.5)},
        anchor=center,
        rotate=-180,
      },
      legend columns=1,
      legend style={
        at={(0.5,1.08)},
        anchor=south,
        cells={anchor=west},
      },
      y axis line style={blue},
      y tick style={blue},
      yticklabel style={text=blue},
    ]
    \addplot+[color=red, mark=o, mark size=3pt]
    coordinates {
    (0,0)
    (0.1, 0)
    };
    \addplot+[color=blue, mark=triangle, mark size=3pt]
    coordinates {
    (1.0, 27.859)
    (3.8, 30.198)
    (14.5, 34.197)
    };
    \legend{Training time, Inference time};

    \end{axis}
    \end{tikzpicture}
    \caption{Training times with full pre-training dataset ($B = 64$) and inference times ($B = 9$)}
    \end{subfigure}
    
    \begin{subfigure}{0.9\linewidth}
    \begin{tikzpicture}
    \begin{axis}[
      width=\textwidth,
      height=0.6\textwidth,
      axis lines*=left,
      xmin=0.9, xmax=15,
      xtick={1, 3.8, 14.5},
      xticklabels={1.0M, 3.8M, 14.5M},
      xlabel={AeroTransformer Sizes},
      ylabel={Peak memory (GB)},
      ylabel style={yshift=-15pt},
    ]
    
    \addplot+[color=black, mark=o, mark size=3pt]
    coordinates {
    (1.0, 9.047)
    (3.8, 12.066)
    (14.5, 18.459)
    };

    \end{axis}
    \end{tikzpicture}
    \caption{Training peak memory ($B = 64$)}
    \end{subfigure}
    \end{minipage}
\caption{AeroTransformer performance when model and dataset size are scaled up}\label{fig:scaleall}
\end{figure}
\end{modifiedblock}

The results demonstrate several trends that are broadly consistent with the expected scaling behavior. First, across all three model sizes, increasing the fraction of pre-training data leads to a clear decrease in test error on the held-out test split. The gains, however, diminish as the dataset grows. For the S-size model, this saturation can be attributed to its limited representational capacity. However, for the larger models, the weaker improvement with more data is more likely due to the fixed training budget, since all configurations share the same number of training steps. Although we did not further increase the number of training steps in this test scenario, the trend in Appendix \ref{app:trainconfig}.\ref{app:steps} suggests that there are still benefits to be gained from additional steps.

Second, with sufficient training data, enlarging the model from S to L reduces both training and test errors, indicating that additional capacity improves the surrogate's expressiveness and its ability to capture complex surface flow patterns. \modified{This improvement, however, comes at the cost of increased computational resources, with both training time and peak memory usage approximately doubling from S to L. Nevertheless, inference time increases only marginally, remaining around 30 ms across all model sizes.}

\subsubsection{Predictions of aerodynamic coefficients}

This section compares the two approaches proposed in this paper for predicting aerodynamic coefficients, which are crucial for downstream shape optimization tasks. 

\paragraph{Avoiding overfitting}

Table \ref{tab:aeroperf} provides a comparison of the two approaches, namely, to directly use coefficients as the primary output, or to use surface flow as the primary output and obtain the coefficients with integration. 
For the former approach, we utilize the coefficient-output version AeroTransformer $\mathrm{AT}_\mathrm{coef}$ introduced in Sec. \ref{sec:aerooutput}. We also tested random forests (RFs) and light gradient boosting machines (LGBMs) to predict coefficients from all shape parameters, rather than using the mesh. Their implementation are provided in Appendix \ref{app:baseline}. The three models for direct coefficient outputs have a similar number of trainable parameters as $\mathrm{AT}_\mathrm{surf}$ at 1.0M. 

\begin{table}[htbp]
    \centering
    \caption{\label{tab:aeroperf}Performance comparison for aerodynamic coefficient prediction between $\mathrm{AT}_\mathrm{surf}$ and $\mathrm{AT}_\mathrm{coefficient}$}
    \begin{tabular}{cccccccc}
        \hline\hline
        \multirow{3}{*}{\makecell{Primary \\ output}} & \multirow{3}{*}{Model} & \multicolumn{6}{c}{Errors} \\ 
        & & \multicolumn{3}{c}{Training set} & \multicolumn{3}{c}{Testing set}\\
        & & $\delta C_L$ & $\delta C_D$ & $\delta C_{M,z}$ & $\delta C_L$ & $\delta C_D$ & $\delta C_{M,z}$  \\
        &  & \mdf ($\times 10^{-3}$) & \mdf ($\times 10^{-4}$) & \mdf ($\times 10^{-3}$)  & \mdf  ($\times 10^{-3}$) & \mdf ($\times 10^{-4}$) & \mdf ($\times 10^{-3}$) \\
        \midrule
        \multirow{3}{*}{Coefficients} & Random Forest & 13.01 & 10.87 & 17.07 & 43.86 & 36.49 & 58.96 \\
        & LGBM & 2.60 & 2.27 & 3.09 & 19.65 & 16.77 & 25.46 \\
        & $\mathrm{AT}_\mathrm{coef}$ (Ours) & 1.02 & 0.80 & 1.11 & 7.19 & 5.98 & 8.45 \\ 
        Surface flow & $\mathrm{AT}_\mathrm{surf}$ (Ours) & 0.83 & 0.84 & 0.90 & 1.31 & 1.19 & 1.55 \\
        \hline\hline
    \end{tabular}
\end{table}

We first evaluate decision-tree-based regression models. Although they are easy to train, their representational capacity is limited for the highly nonlinear problem in this work. We also tested increasing the model size by increasing the number of estimators, but this yielded only marginal improvements. Moreover, these methods rely on specific geometric parameterizations, which limit their applicability to more complex task-specific fine-tuning datasets.

The $\mathrm{AT}_\mathrm{coef}$ takes mesh-based geometry as input and exhibits stronger expressive power. It achieves training-set performance comparable to $\mathrm{AT}_\mathrm{surf}$, but all variants that use coefficients as the primary output show varying degrees of overfitting. The $\mathrm{AT}_\mathrm{coef}$, in particular, exhibits testing errors more than five times larger. In contrast, the model with surface flow as the primary output significantly alleviates overfitting. The surface flow acts as a regularization term when used as the primary output. Additionally, because lift and drag differ substantially in magnitude, even careful normalization fails to guarantee comparable error scales across the quantities. In contrast, the integration step to get coefficients from $\mathrm{AT}_\mathrm{surf}$ naturally enforces consistent error magnitudes for $C_L$ and $C_D$. 

Overall, these results demonstrate that a better approach to predict the coefficients is to use the surface flow as the primary output and calculate the aerodynamic coefficients from it.

\paragraph{Extra aerodynamic coefficient loss}

Since using surface flow as a primary output has been shown to yield better coefficient prediction, we further attempted to enhance coefficient accuracy by introducing an aerodynamic loss term into $\mathrm{AT}_\mathrm{surf}$. We tested the sum of the two loss terms in equation \ref{eqn:loss} with different weights and found that $\lambda = 0.1$ leads to a similar magnitude of the two loss terms after convergence. In addition to this value, we also provided results for larger and smaller $\lambda$ values: $0.5$ and $0.05$. The results of updating the gradient with the ConFIG are also provided. The values shown in Table \ref{tab:aeroloss} represent the errors on the testing sets, averaged across three cross-validation runs.

\begin{table}[htbp]
    \centering    \caption{\label{tab:aeroloss}$\mathrm{AT}_\mathrm{surf}$ performance with and without aerodynamic loss term}
    \begin{tabular}{cccccccc}
        \hline\hline
        \multicolumn{2}{c}{\multirow{2}{*}{Model}} & \multicolumn{6}{c}{Errors} \\ 
        && $\delta C_p$ & $\delta C_{f,\tau}$ & $\delta C_{f,z}$ & $\delta C_L$ & $\delta C_D$ & $\delta C_{M,z}$ \\
        && \multicolumn{3}{c}{\mdf Percentage relative error (\%)} & \mdf ($\times 10^{-3}$) & \mdf ($\times 10^{-4}$) & \mdf ($\times 10^{-3}$) \\
        \midrule
        \multicolumn{2}{c}{w/o aero. loss} & 0.157 & 0.120 & 0.147 &  1.31 & 1.19 & 1.55 \\
        \multirow{3}{*}{\makecell{w/ aero. loss\\(weighted)}} &\mdf  $\lambda=0.05$ &\mdf  0.158 &\mdf  0.121 &\mdf  0.149 &\mdf   1.20 & \mdf \textbf{1.15} &\mdf  1.43\\
        & $\lambda=0.1$ & \textbf{0.154} & \textbf{0.119} & \textbf{0.146} &  1.11 & 1.17 & \textbf{1.33}  \\
        &\mdf  $\lambda=0.5$ &\mdf  0.165 &\mdf  0.127 & \mdf 0.154 &\mdf   \textbf{1.09} &\mdf  1.21 &\mdf  1.34  \\
        \multicolumn{2}{c}{w/ aero. loss (ConFIG)} & 0.216 & 0.162 & 0.195 &  1.51 & 1.53 & 1.84 \\
        \hline\hline
    \end{tabular}
\end{table}

\modified{Introducing the aerodynamic loss term generally improves the prediction of aerodynamic coefficients compared to the baseline without this term. Yet, different values of $\lambda$ result in only slightly different values, and for the three coefficients considered, they achieve their optimal values with different $\lambda$.} Surprisingly, however, the inclusion of the aerodynamic loss does not degrade surface flow accuracy. In fact, for $\lambda = 0.1$, the surface flow error becomes lower than the baseline model trained without any aerodynamic loss. We tested this behavior across different dataset sizes and training configurations and found that it consistently occurs. This suggests that aerodynamic losses not only improve the accuracy of the integrated aerodynamic coefficients but also act as a global regularizer, potentially suppressing spatially inconsistent artifacts and ultimately leading to improved surface flow predictions. Therefore, incorporating an aerodynamic loss term during training is broadly beneficial. Another interesting behavior is that training with $\lambda=0.1$ yields a lower surface flow error, whereas $\lambda=0.5$ achieves slightly better accuracy in predicting the lift coefficient $C_L$. This behavior can also be observed in Table \ref{tab:aeroperf} between ViT and Transolver. A detailed examination of the error distribution is provided in the Appendix \ref{app:mismatch}.

Nonetheless, achieving optimal performance requires tuning the weight $\lambda$, as excessively large or small values reduce the effectiveness of the added supervision. 
This motivates the use of the parameter-free gradient ConFIG method. However, it does not achieve the level of accuracy obtained with manual choices of $\lambda$, most likely because the gradients are predominantly aligned, which reduces the advantages of the conflict-resolving optimizer.

\subsection{Task-specific application}

The results in the above section are based on data drawn from the same distribution as the pre-training dataset. In practical applications, performance on downstream tasks with different geometric distributions is potentially even more important. This section examines the fine-tuning of the model on a task-specific downstream dataset generated by perturbing a baseline CRM wing.

\subsubsection{Advantages of using the pre-trained model}

We first show the benefits of building a task-specific local surrogate model based on the pre-trained model, compared to training it from scratch. We use 20\% of the fine-tuning dataset in Table \ref{tab:datasets}. This corresponds to approximately 450 samples (58 wing shapes), generated in 30.4 hours using a 160-processor CPU cluster. \modified{This cost is comparable to a single adjoint-based optimization run} \cite{kenway_multipoint_2016}. Given the dataset's relatively small size, we use 10-fold cross-validation. For each of the ten runs, after 10\% of the testing samples are set aside, we randomly select 20\% of the remaining training samples to train the model, then test its performance on the entire testing set.

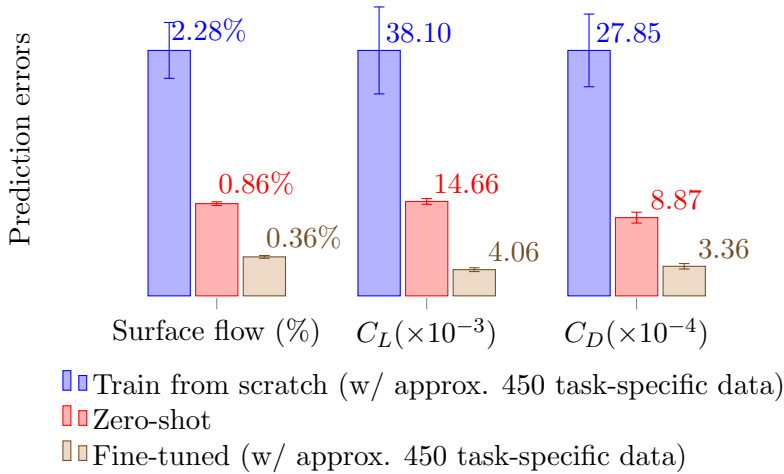
\begin{figure}[ht]
\centering
\footnotesize
\begin{tikzpicture}
\begin{axis}[
  ybar,
  width=0.6\textwidth,
  height=0.35\textwidth,
  bar width=16pt,
  enlarge x limits=0.25,
  xtick={1,4,7},
  xticklabels={Surface flow (\%), $C_L(\times 10^{-3})$, $C_D(\times 10^{-4})$},
  axis x line*=bottom,
  ytick=\empty,
  ylabel={Prediction errors},
  axis line style={draw=none},
  point meta=explicit,
nodes near coords={
    \pgfmathprintnumber[fixed,fixed zerofill,precision=2]{\pgfplotspointmeta}\ifnum\coordindex=0\relax\%\fi
},
  nodes near coords style={xshift=15pt},
  legend style={
    at={(0.5, -0.2)},
    anchor=north,
    legend columns=1,
    cells={anchor=west},
    draw=none
  }
]

\addplot+[xshift=0pt, error bars/.cd, y dir=both,y explicit] 
coordinates {
(1,1.00000) +- (0,0.11369) [2.28]
(4,1.00000) +- (0,0.17714) [38.10]
(7,1.00000) +- (0,0.14808) [27.85]
};

\addplot+[xshift=0pt, error bars/.cd, y dir=both,y explicit] 
coordinates {
(1,0.37575) +- (0,0.00744) [0.86]
(4,0.38471) +- (0,0.01164) [14.66]
(7,0.31867) +- (0,0.02209) [8.87]
};

\addplot+[xshift=0pt, error bars/.cd, y dir=both,y explicit] 
coordinates {
(1,0.15882) +- (0,0.00563) [0.36]
(4,0.10667) +- (0,0.00772) [4.06]
(7,0.12049) +- (0,0.01108) [3.36]
};

\legend{Train from scratch (w/ approx. 450 task-specific data), Zero-shot, Fine-tuned (w/ approx. 450 task-specific data)}

\end{axis}
\end{tikzpicture}

\caption{Impact of pre-training on task-specific prediction errors}
\label{fig:transfer}
\end{figure}

We compare the mean error on the ten runs in Fig. \ref{fig:transfer} for training an L-size AeroTransformer from scratch and for fine-tuning it with weights initialized from the best pre-trained model on the full pre-training dataset. We also include the average zero-shot performance across the entire task-specific dataset for the L-size AeroTransformer pre-trained on the full pre-training dataset. For both from-scratch training and fine-tuning, all model parameters are trainable, and the number of training steps is 5.6k, resulting in a training time of 49.6 minutes. \modified{We note that the cost of generating a task-specific dataset and fine-tuning the model varies with the budget, and that the resulting models generally improve as the budget increases. We will discuss this in Sec.} \ref{sec:finetunetime}.

\begin{figure}[htbp]
    \centering
    \begin{subfigure}{1\textwidth}
        \centering
        \includegraphics[width=1\linewidth]{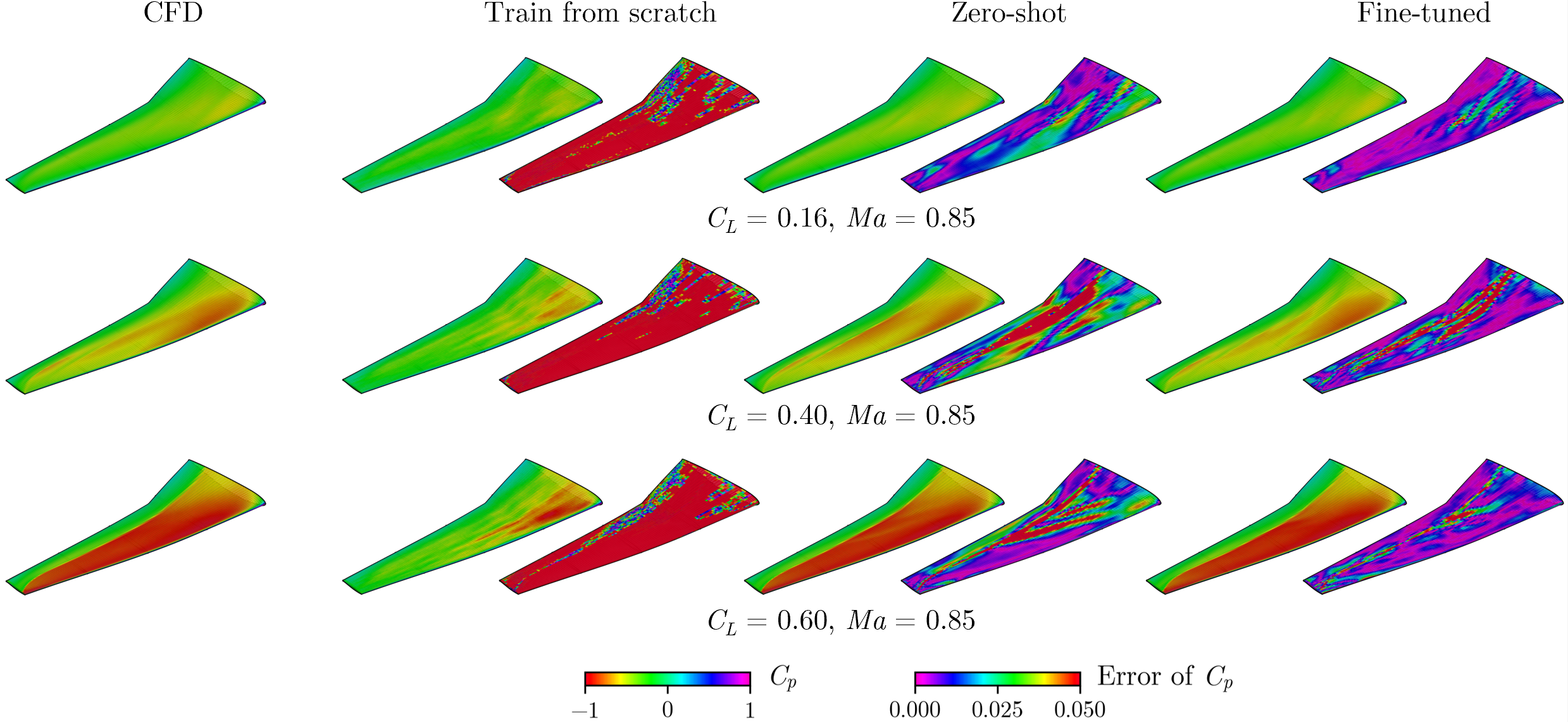}
        \caption{Surface pressure coefficient contours}
        \label{fig:crmcp}
    \end{subfigure}
    \begin{subfigure}{1\textwidth}
        \centering
        \includegraphics[width=0.9\linewidth]{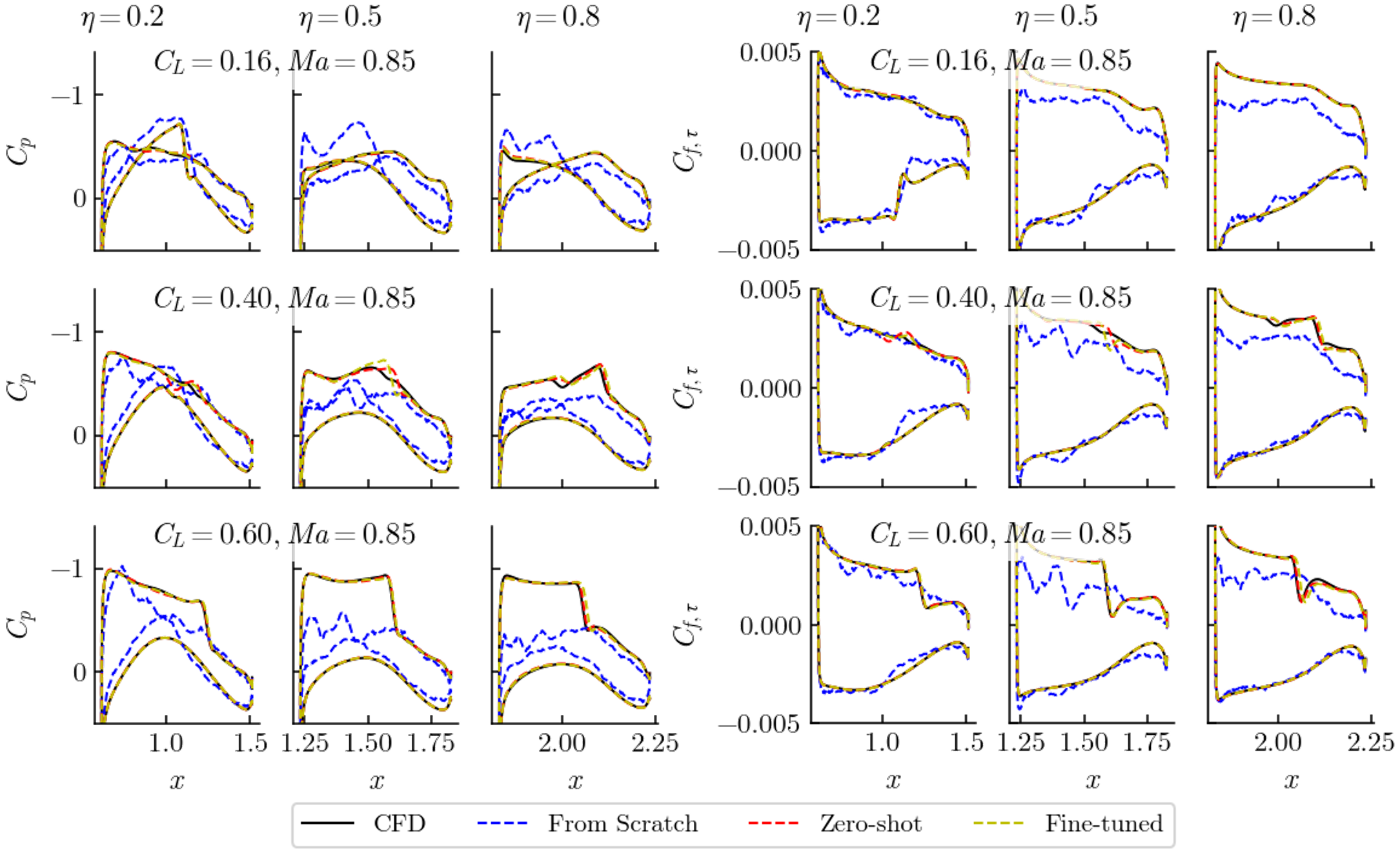}
        \caption{Cross-sectional coefficients distribution}
        \label{fig:testwingeta}
    \end{subfigure}

    \caption{Impact of pre-training on the surface flow prediction performance of CRM}
    \label{fig:crmtest}
\end{figure}

\begin{figure}[ht]
    \centering
    \includegraphics[width=0.75\linewidth]{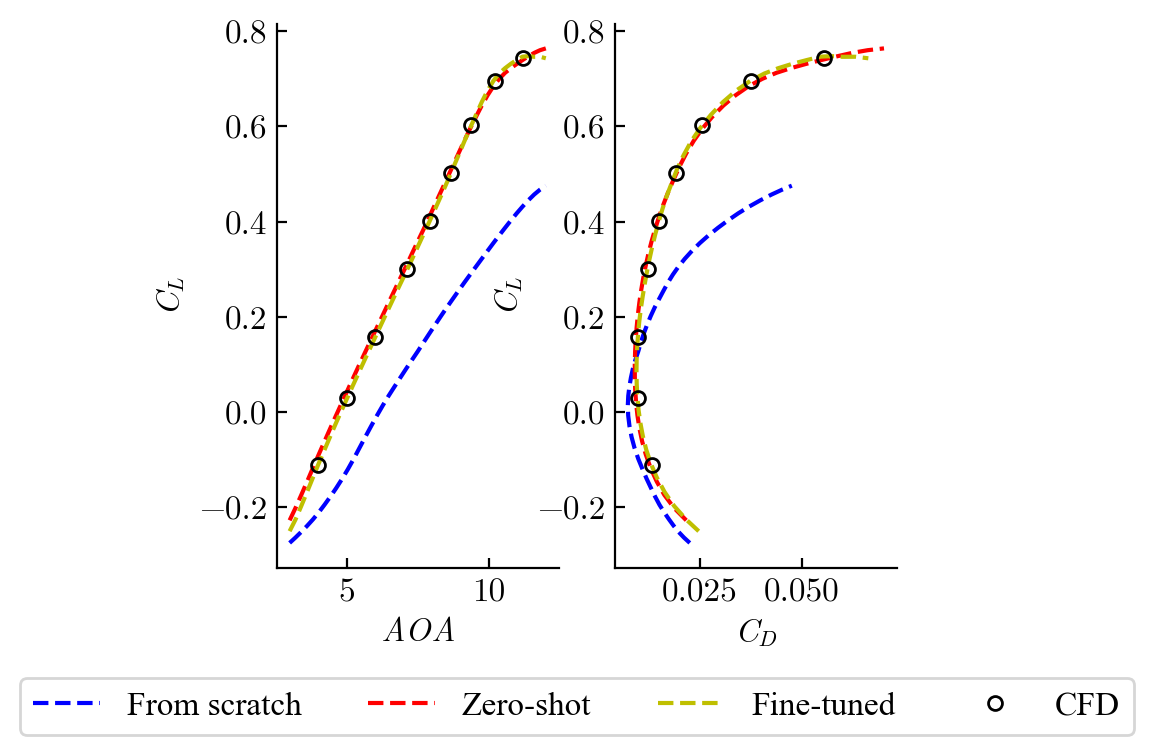}
    \caption{Impact of pre-training on the aerodynamic curves of CRM}
    \label{fig:crmcoef}
\end{figure}

We also provide an intuitive demonstration of the model's performance on the unperturbed CRM wing, which serves as the baseline wing shape for, but is not included in the task-specific dataset. In Fig. \ref{fig:crmtest}, the surface pressure coefficient contours under Mach number 0.85 and three lift coefficients are provided in Fig. \ref{fig:crmtest}(a). For each method, the left wings show the pressure coefficient contours, while the right ones show the relative error. Figure \ref{fig:crmtest}(b) provides the sectional distributions of pressure and streamwise friction coefficients on three spanwise stations at $\eta$ = 0.2, 0.5, and 0.8. Figure \ref{fig:crmcoef} provides the lift curve and polar curve of the CRM wing at Mach number 0.85.   

\paragraph{Zero-shot performance} The comparisons show that, even though the model was pre-trained on a dataset with relatively limited details (a single baseline sectional airfoil shape is used), the zero-shot performance of the model is surprisingly good on a quite different task-specific dataset (each wing has seven different sectional airfoils). Compared to training a model from scratch, the zero-shot model not only eliminates task-specific training consumption but also reduces the error by approximately 64.0\%. The intuitive illustrations on the CRM wing further strongly demonstrate the accuracy of the zero-shot pre-trained model, which effectively captures the flow structures and crucial aerodynamic coefficients. In contrast, the model trained from scratch performs poorly.

\paragraph{Fine-tuning performance} Fine-tuning the pre-trained model \modified{with about 450 samples} can further benefit the model performance on the task-specific dataset. This is most obvious in Fig. \ref{fig:trans}, where the prediction errors are reduced by nearly 84.2\% compared to without a pre-trained model. When compared to zero-shot performance, it also reduces 2/3 of the prediction error. The advantage is less obvious on the surface flow and aerodynamic curves; nevertheless, the improvement under operating conditions that are away from the cruise condition can still be observed in Fig. \ref{fig:crmcoef}.

Together, these results clearly demonstrate that pre-training enables the model to capture the essential relationship between geometry and canonical flow patterns within a reduced yet expressive geometric representation. Consequently, a small amount of or even no task-specific data is sufficient to adapt the pre-trained model to achieve accurate predictions on more complex shape configurations.

\subsubsection{Choosing the pre-trained model}

\modified{An important observation in the foundation-model paradigm is that a stronger base model consistently leads to improved downstream performance} \cite{kaplan_scaling_2020}, \modified{which motivates scaling up the pre-training stage. To examine whether this trend holds for our model and methodology, we tested several pre-trained models with smaller parameter counts and reduced training data for their downstream performance. The fine-tuning settings are kept identical across all base models to ensure a fair comparison.

In Fig.}~\ref{fig:transsize}, \modified{we report the surface flow error on the fine-tuning dataset. The $x$-axis represents the relative scale with respect to the strongest model that we analyzed in the last section. This normalization enables direct comparison of scaling effects across model and dataset sizes.}

\begin{modifiedblock}
\begin{figure}[H]
\centering
\small
\begin{tikzpicture}
\begin{axis}[
  width=0.6\textwidth,height=0.4\textwidth,
  axis lines*=left,
  xmin=0.08,xmax=1.02,
  xmode=log,
  xlabel={Scale percentage w.r.t the strongest base model},
  xlabel style={at={(axis description cs:0.5,0.04)},anchor=north},
  xtick={0.2, 0.4, 0.6, 0.8, 1.0},
  xticklabels={0.2, 0.4, 0.6, 0.8, 1.0},
  ylabel={Surface flow error ($SFE$)},
  yticklabel={
        \pgfmathparse{\tick}
        \pgfmathprintnumber[fixed,fixed zerofill,precision=2]{\pgfmathresult}\%
    },
]

\addplot+[color=red, mark=o, mark size=3pt,
  nodes near coords={ \pgfmathprintnumber[fixed,fixed zerofill,precision=3]{\pgfplotspointmeta}\%},
  nodes near coords style={xshift=12pt,},
  error bars/.cd, y dir=both,y explicit,]
coordinates {
(0.0952,0.36951) +- (0,0.01210)
(0.362,0.36432) +- (0,0.01107)
(1.0,0.36150) +- (0,0.01282)
};

\addplot+[color=blue, dashed, mark=o, mark size=3pt,
  nodes near coords={ \pgfmathprintnumber[fixed,fixed zerofill,precision=3]{\pgfplotspointmeta}\%},
  nodes near coords style={xshift=12pt,},
  error bars/.cd, y dir=both,y explicit,]
coordinates {
(0.11111, 0.60632) +- (0, 0.01204)
(0.27778, 0.48274) +- (0, 0.02487)
(0.55556, 0.39913) +- (0, 0.01890)
(0.83333, 0.37701) +- (0, 0.01617)
(1.00000, 0.36150) +- (0, 0.01329)
};

\legend{varying pre-trained model size, varying pre-training dataset size}

\end{axis}

\begin{axis}[
  anchor=south west, yshift=-30pt,
  width=0.6\textwidth,height=0.4\textwidth,
  axis x line*=bottom,
  axis y line*=none,
  xmin=0.08,xmax=1.02,
  ymin=0, ymax=10,
  xmode=log,
  xlabel={Pre-trained model size},
  xlabel style={red,at={(axis description cs:0.5,0.04)},anchor=north},
  xtick={0.0952, 0.362, 1.0},
  xticklabels={1.0M, 3.8M, 14.5M},
  ytick=\empty,
  clip=false,
  background/.style={},
  x axis line style={red},
  x tick style={red},
  xticklabel style={text=red},
]
\end{axis}

\begin{axis}[
  anchor=south west, yshift=-60pt,
  width=0.6\textwidth,height=0.4\textwidth,
  axis x line*=bottom,
  axis y line*=none,
  xmin=0.08,xmax=1.02,
  ymin=0, ymax=10,
  xmode=log,
  xlabel={Pre-training dataset size},
  xlabel style={blue,at={(axis description cs:0.5,0.04)},anchor=north},
  xtick={0.11111, 0.27778, 0.55556, 0.83333, 1.00000},
  xticklabels={2.6k, 6.5k, 12.9k,, 23.4k},
  ytick=\empty,
  clip=false,
  background/.style={},
  x axis line style={blue},
  x tick style={blue},
  xticklabel style={text=blue},
]
\end{axis}
\end{tikzpicture}
\caption{Downstream performance with different pre-trained models}\label{fig:transsize}
\end{figure}
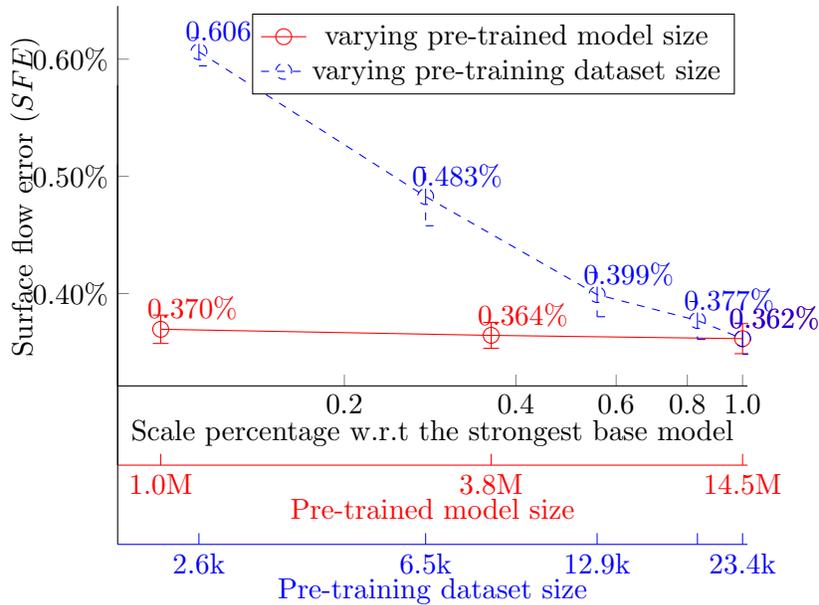
\end{modifiedblock}

\modified{The results show that downstream performance benefits from a stronger pre-trained model, either larger model size or larger pre-training dataset size. Within the range considered, the dataset size has a more pronounced impact on performance; however, expanding the dataset typically incurs significantly higher computational cost.}

\subsubsection{Task-specific budget for data generation and model fine-tuning}\label{sec:finetunetime}

\modified{In the previous sections, we considered a moderate task-specific budget, where approximately 450 flow field samples were generated and the model was fine-tuned for 5.6k steps. Compared to the pre-training stage, fine-tuning relies on a much smaller number of samples. Given sufficient computational resources, the wall-clock time for data generation can be comparable to that of model fine-tuning, making the latter a significant component of the overall computational budget. In practical applications, however, the available computational budget may vary substantially across use cases. To better understand the trade-offs involved, we analyze in the following how the size of the fine-tuning dataset and the corresponding training configurations influence model performance. }

\paragraph{Influence of downstream data size}

\modified{We fine-tune the L-size pre-trained model with a minimum of 29 to a maximum of 2160 samples. We keep using the ten-fold cross-validation scheme and consider two strategies to construct the training dataset from the samples outside the held-out fold in each run:}

\begin{itemize}
    \item \modified{Selecting a subset of shapes and including all corresponding flow fields under different operating conditions.}
    \item \modified{First selecting a subset of shapes and then randomly selecting a single flow field for shapes.}
\end{itemize}

Figure~\ref{fig:finetunesize} \modified{shows the surface flow prediction errors of models fine-tuned with different amounts of training data. For comparison, we also train models from scratch with the same number of samples and include the pre-trained model's zero-shot performance as the leftmost reference point. Since our goal is to scale up task-specific investment, we scale the training steps linearly with the dataset size while keeping the number of training epochs constant. }

\begin{modifiedblock}
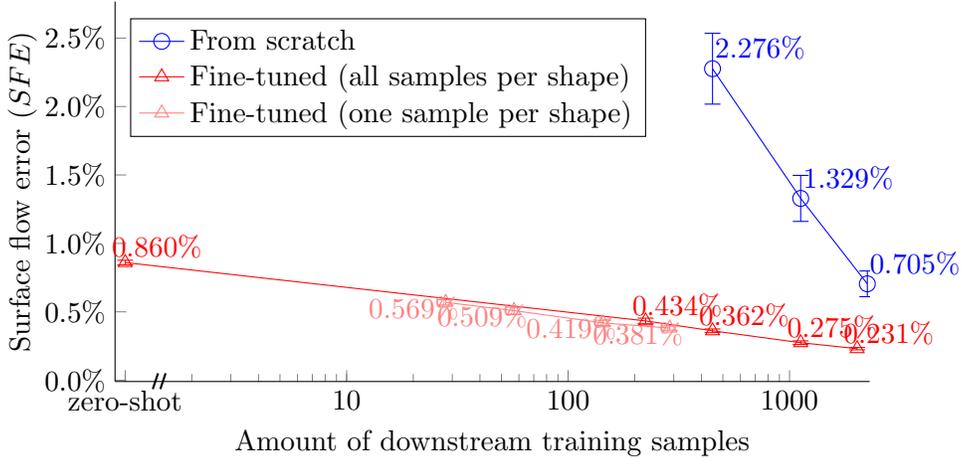
\begin{figure}[H]
\centering
\small
\begin{tikzpicture}
\begin{axis}[
  width=0.7\textwidth,
  height=0.4\textwidth,
  axis lines*=left,
  xmin=0.9,xmax=2300,
  xlabel={Amount of downstream training samples},
  xmode=log,
  xtick={1, 10, 100, 1000},
  xticklabels={zero-shot, 10, 100, 1000},
  ylabel={Surface flow error ($SFE$)},
  yticklabel={
        \pgfmathparse{\tick}
        \pgfmathprintnumber[fixed,fixed zerofill,precision=1]{\pgfmathresult}\%
    },
  legend columns=1,
  legend style={
    at={(0.02,0.8)},
    anchor=west,
    cells={anchor=west},
  },
  after end axis/.code={
    \draw[black, line width=0.8pt]
      (rel axis cs:0.05,-0.02) -- ++(2pt,6pt)
      (rel axis cs:0.06,-0.02) -- ++(2pt,6pt);
  },
]

\addplot+[
  color=blue,
  mark=o,
  mark size=3pt,
  nodes near coords={ \pgfmathprintnumber[fixed,fixed zerofill,precision=3]{\pgfplotspointmeta}\%},
  nodes near coords style={xshift=18pt,},
  error bars/.cd, y dir=both,y explicit,
]
coordinates {
(449, 2.27618) +- (0, 0.25878)
(1123, 1.32870) +- (0, 0.16808)
(2246, 0.70457) +- (0, 0.09388)
};

\addplot+[
  color=red,
  mark=triangle,
  mark size=3pt,
  nodes near coords={ \pgfmathprintnumber[fixed,fixed zerofill,precision=3]{\pgfplotspointmeta}\%},
  nodes near coords style={xshift=12pt,yshift=-2pt,},
  error bars/.cd, y dir=both,y explicit,
]
coordinates {
(1, 0.86) +- (0, 0.017)
(224.00000, 0.43366) +- (0, 0.02038)
(449.00000, 0.36150) +- (0, 0.01329)
(1123.00000, 0.27458) +- (0, 0.01322)
(2021.00000, 0.23093) +- (0, 0.00866)
};

\addplot+[
  color=red!50,
  mark=triangle,
  mark size=3pt,
  nodes near coords={ \pgfmathprintnumber[fixed,fixed zerofill,precision=3]{\pgfplotspointmeta}\%},
  nodes near coords style={xshift=-12pt,yshift=-10pt,},
  error bars/.cd, y dir=both,y explicit,
]
coordinates {
(28.00000, 0.56928) +- (0, 0.01939)
(57.00000, 0.50861) +- (0, 0.02509)
(144.00000, 0.41923) +- (0, 0.01935)
(288.00000, 0.38126) +- (0, 0.01591)
};

\legend{From scratch, Fine-tuned (all samples per shape), Fine-tuned (one sample per shape)};

\end{axis}
\end{tikzpicture}
\caption{Model performance trained with different amounts of task-specific samples}
\label{fig:finetunesize}
\end{figure}
\end{modifiedblock}

\modified{When varying the amount of task-specific training samples, we observe a near-linear improvement in downstream performance as the dataset size increases. Notably, even with $\mathcal{O}(10)$ samples, the model achieves a surface flow prediction error of approximately 0.5\% (i.e., about 4.5 counts in $C_D$). With a budget of $\mathcal{O}(10^3)$ samples, the error is further reduced to 0.231\% (i.e., about 1.7 counts in $C_D$).

For relatively small datasets, fine-tuning with one sample per shape consistently outperforms using fewer shapes with multiple operating conditions per shape. This suggests that maintaining diversity in the geometric space is more important than in the direction of operating conditions for the task-specific dataset. We attribute this behavior to the pre-trained model's stronger ability to generalize across operating conditions.

Compared with models trained from scratch, the fine-tuning approach provides a significant advantage across all data budgets. Although the error reduction slope of the from-scratch model is steeper, extrapolation suggests that it would require $\mathcal{O}(10^4)$ samples to match the performance of the fine-tuned model, which is typically impractical for downstream applications. 

These results demonstrate that building task-specific surrogate models on top of a pre-trained model substantially improves data efficiency and enables effective performance even under limited task-specific computational budgets.}

\paragraph{Influence of fine-tuning steps} 

\modified{In this paragraph, we fix the fine-tuning dataset to 450 samples, and vary the number of fine-tuning steps for both the pre-trained model and the model trained from scratch to study its effect. The results are shown in} Fig.~\ref{fig:finetunestep}.

\modified{We observe that increasing the number of training steps leads to error reduction for both models. However, the rate of improvement is similar in both cases, indicating the aforementioned advantage of the fine-tuned model across a wide range of training settings. More importantly, both results exhibit clear saturation behavior. Even with a substantially larger optimization budget, the model trained from scratch saturates at around 1.7\% error, which is still well above that of the fine-tuned model. This observation suggests that the fine-tuned model's better performance is not due to a better starting point, but rather to the transferable knowledge learned during pre-training.} 

\begin{modifiedblock}
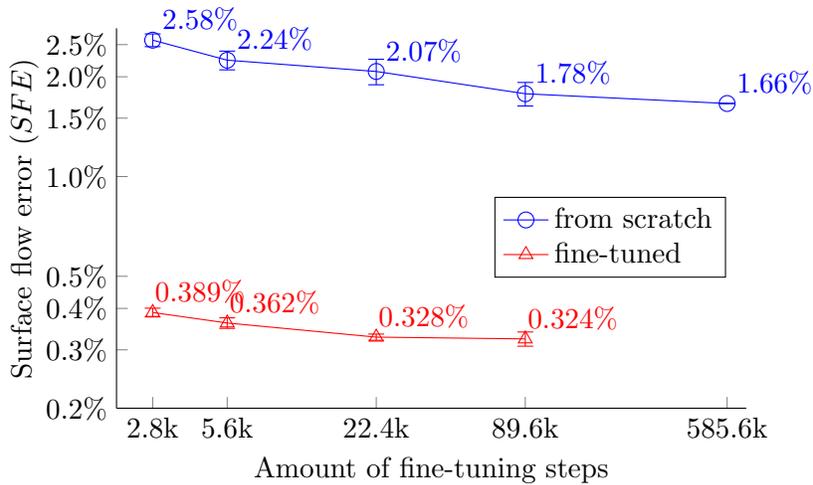
\begin{figure}[H]
\centering
\small
\begin{tikzpicture}
\begin{axis}[
  width=0.6\textwidth,
  height=0.4\textwidth,
  axis lines*=left,
  xmin=2,xmax=700,
  ymin=0.2,ymax=2.8,
  xlabel={Amount of fine-tuning steps},
  ylabel={Surface flow error ($SFE$)},
  xtick={2.8, 5.6, 22.4, 89.6, 585.6},
  xticklabels={2.8k, 5.6k, 22.4k, 89.6k, 585.6k},
  xmode=log,
  ymode=log,
  ytick={0.2, 0.3, 0.4, 0.5, 1, 1.5, 2, 2.5},
  yticklabels={0.2\%, 0.3\%, 0.4\%, 0.5\%, 1.0\%, 1.5\%, 2.0\%, 2.5\%},
  legend columns=1,
  legend style={at={(0.6,0.45)},anchor=west,cells={anchor=west}},
]
\addplot+[color=blue, mark=o, mark size=3pt,
  point meta=explicit,
  nodes near coords={ \pgfmathprintnumber[fixed,fixed zerofill,precision=2]{\pgfplotspointmeta}\%},
  nodes near coords style={xshift=18pt,},
  error bars/.cd, y dir=both,y explicit,]
coordinates {
(2.80000, 2.57918) +- (0, 0.11407) [2.57918]
(5.60000, 2.24418) +- (0, 0.14583) [2.24418]
(22.40000, 2.07479) +- (0, 0.18298) [2.07479]
(89.60000, 1.77729) +- (0, 0.14500) [1.77729]
(585.60000, 1.6611) +- (0, 0.00) [1.6611]
};

\addplot+[color=red,mark=triangle,mark size=3pt,
  point meta=explicit,
  nodes near coords={ \pgfmathprintnumber[fixed,fixed zerofill,precision=3]{\pgfplotspointmeta}\%},
  nodes near coords style={xshift=18pt,},
  error bars/.cd, y dir=both,y explicit,]
coordinates {
(2.80000, 0.38887) +- (0, 0.01178) [0.38887]
(5.60000, 0.36150) +- (0, 0.01329) [0.36150]
(22.40000, 0.32770) +- (0, 0.00709) [0.32770]
(89.60000, 0.32367) +- (0, 0.01594) [0.32367]
};
\legend{from scratch, fine-tuned}

\end{axis}
\end{tikzpicture}
\caption{Model performance fine-tuned with increasing training steps}
\label{fig:finetunestep}
\end{figure}
\end{modifiedblock}

\paragraph{Balancing data generation and model fine-tuning}

\modified{In the previous sections, we showed that both increasing the size of the fine-tuning dataset and increasing the number of training steps can improve downstream performance, at the cost of additional computational time. Here, we further analyze how to allocate this time budget effectively.

To make the analysis generalizable across different computational environments, we introduce a time-consumption ratio, $\gamma$, defined as the cost of generating 10 samples relative to 1k training steps. A larger $\gamma$ indicates that data generation is more expensive compared to training, while a smaller $\gamma$ corresponds to scenarios where data can be generated efficiently through parallel computation. 
For the hardware used in this work (2 nodes, each with 80 Intel Xeon Gold 5320 CPU cores), we obtain $\gamma \approx 4.6$. With increased parallel resources, this ratio can be significantly reduced. For example, using a 16-node cluster with the same configuration reduces the ratio to $\gamma \approx 0.57$.

We then use the 450 samples and 5.6k steps as a reference, and display the results from the previous sections in }Fig.~\ref{fig:finetunebalance}, \modified{where performance is plotted against the additional time spent on either increasing dataset size or increasing training steps. The results show that the optimal strategy depends strongly on $\gamma$. When $\gamma$ is large, allocating additional budget to training steps yields more efficient performance improvements and vice versa.}

\begin{modifiedblock}
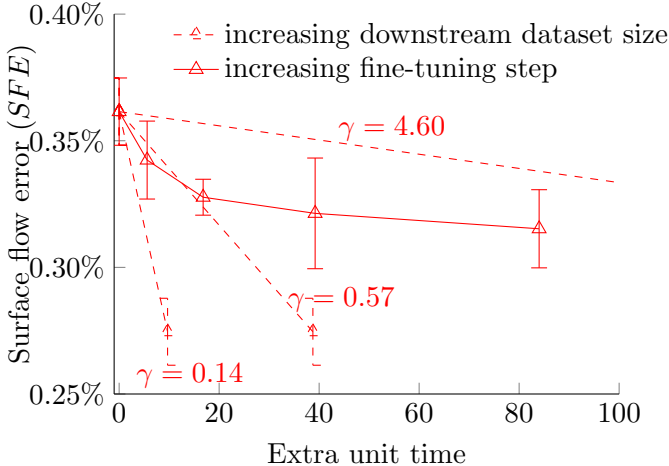
\begin{figure}[H]
\centering
\small
\begin{tikzpicture}
\begin{axis}[
  width=0.5\textwidth,
  height=0.4\textwidth,
  axis lines*=left,
  xmin=-1,xmax=100,
  ymin=0.25,ymax=0.4,
  xlabel={Extra unit time},
  ylabel={Surface flow error ($SFE$)},
  ytick={0.25, 0.3, 0.35, 0.4},
  yticklabels={0.25\%, 0.30\%, 0.35\%, 0.40\%},
  legend columns=1,
  legend style={at={(0.1,0.9)},anchor=west,cells={anchor=west},draw=none},
]

\addplot+[color=red,dashed,mark=triangle,mark size=3pt,
  error bars/.cd, y dir=both,y explicit,]
coordinates {
(0.00000, 0.36150) +- (0, 0.01329)
(310.00320, 0.27458) +- (0, 0.01322)
(723.34080, 0.23093) +- (0, 0.00866)
};

\addplot+[color=red,dashed,mark=triangle,mark size=3pt,
  error bars/.cd, y dir=both,y explicit,]
coordinates {
(0.00000, 0.36150) +- (0, 0.01329)
(9.68760, 0.27458) +- (0, 0.01322)
};

\addplot+[color=red,dashed,mark=triangle,mark size=3pt,
  error bars/.cd, y dir=both,y explicit,]
coordinates {
(0.00000, 0.36150) +- (0, 0.01329)
(38.75040, 0.27458) +- (0, 0.01322)
};

\addplot+[color=red,mark=triangle,mark size=3pt,
  error bars/.cd, y dir=both,y explicit,]
coordinates {
(0.00000, 0.36150) +- (0, 0.01329)
(5.60000, 0.34237) +- (0, 0.01539)
(16.80000, 0.32770) +- (0, 0.00709)
(39.20000, 0.32132) +- (0, 0.02185)
(84.00000, 0.31524) +- (0, 0.01543)
};
\node[red] at (rel axis cs:0.55,0.7) {$\gamma = 4.60$};
\node[red] at (rel axis cs:0.45,0.25) {$\gamma = 0.57$};
\node[red] at (rel axis cs:0.15,0.05) {$\gamma = 0.14$};

\legend{increasing downstream dataset size, , ,increasing fine-tuning step}

\end{axis}
\end{tikzpicture}
\caption{Model performance improvement by spending extra time on increasing dataset size and fine-tuning steps}
\label{fig:finetunebalance}
\end{figure}
\end{modifiedblock}

\subsubsection{Parameter-efficient fine-tuning}

To further reduce adaptation cost for downstream aerodynamic tasks, we evaluated two parameter-efficient strategies: (i) updating only the attention layers of the pre-trained AeroTransformer, and (ii) applying LoRA to these layers with different rank values 4 and 16. One benefit of fine-tuning partial parameters is that it prevents instability, thereby enabling a larger learning rate. Specifically, the initial and maximum learning rate in the one-cycle schedule are raised by one order of magnitude for the parameter-efficient strategies.

The results are presented in Table \ref{tab:lora}, all obtained via 10-fold cross-validation on the fine-tuning dataset \modified{with 450 samples}. They reveal a trade-off between fine-tuning cost and accuracy. Not surprisingly, fine-tuning all model parameters yields the best performance, while restricting updates to the attention layers reduces the tunable parameter count to just 15.0\% of full fine-tuning, but only leads to 6.1\% larger prediction errors in the surface flow. For lift and pitching moment coefficients, this parameter-efficient approach even shows better performance. 

\modified{We also provide in} Fig. \ref{fig:pefinetunesize} \modified{a comparison of full-parameter fine-tuning and only attention-layer fine-tuning across different training data. When dealing with very limited samples (e.g., $\mathcal{O}(10^1)$ to $\mathcal{O}(10^2)$), the parameter-efficient approach achieves better performance in surface flow prediction. This can be attributed to its reduced risk of overfitting, as fewer trainable parameters need to be adapted to the small dataset. As the dataset size increases, the advantage of parameter-efficient fine-tuning gradually diminishes.}

Incorporating LoRA further decreases the number of trainable parameters to 0.4\%, but this comes at the cost of progressively higher prediction errors. This indicates that the downstream aerodynamic prediction task requires sufficient expressive capacity beyond LoRA. Overall, these results demonstrate that parameter-efficient fine-tuning provides a flexible spectrum of adaptation strategies with varying cost–performance trade-offs.

\begin{table}[htbp]
    \centering
    \caption{\label{tab:lora}Parameter-efficient fine-tuning performance comparison}
    \begin{tabular}{ccccccccc}
        \hline\hline
        \multicolumn{2}{c}{\multirow{3}{*}{\makecell{Fine-tuning \\ strategy}}} &\multicolumn{6}{c}{Errors} & \multirow{2}{*}{\makecell{\# of tunable param. \\ (\% of all param.)}} \\ 
        & &  $\delta C_p$ & $\delta C_{f,\tau}$ & $\delta C_{f,z}$ & $\delta C_L$ & $\delta C_D$ & $\delta C_{M,z}$ &  \\
        & & \multicolumn{3}{c}{\mdf Percentage relative error (\%)} & \mdf ($\times 10^{-3}$) & \mdf ($\times 10^{-4}$) & \mdf ($\times 10^{-3}$) \\
        \midrule
        \multicolumn{2}{c}{All parameters} & \textbf{0.435} & \textbf{0.302} & \textbf{0.348} & 4.06 & \textbf{3.36} & 4.94 & 14.52 M (100\%) \\ 
        \multirow{3}{*}{\makecell{Only \\ Attn.}} & w/o LoRA & 0.465 & 0.318 & 0.369 & \textbf{3.98} & 3.45 & \textbf{4.89} & 2.19 M (15.0\%) \\ 
        & LoRA16 & 0.499 & 0.347 & 0.407 & 4.14 & 3.66 & 5.03 & 0.24 M (1.7\%) \\ 
        & LoRA4 & 0.567 & 0.389 & 0.462 & 4.50 & 4.15 & 5.57 & 0.06 M (0.4\%) \\ 
        \hline\hline
    \end{tabular}
\end{table}

\begin{modifiedblock}
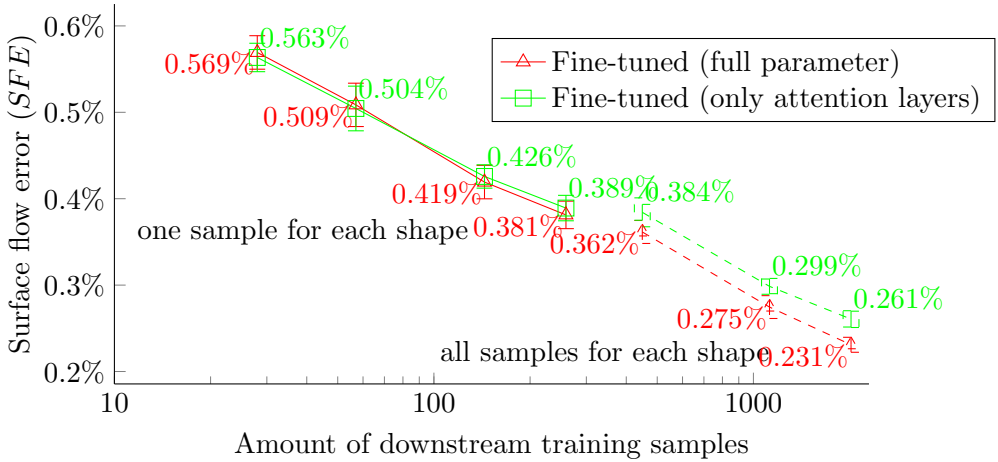
\begin{figure}[H]
\centering
\small
\begin{tikzpicture}
\begin{axis}[
  width=0.7\textwidth,
  height=0.4\textwidth,
  axis lines*=left,
  xmin=10,xmax=2300,
  xlabel={Amount of downstream training samples},
  xmode=log,
  xtick={10, 100, 1000},
  xticklabels={10, 100, 1000},
  ylabel={Surface flow error ($SFE$)},
  yticklabel={
        \pgfmathparse{\tick}
        \pgfmathprintnumber[fixed,fixed zerofill,precision=1]{\pgfmathresult}\%
    },
  legend columns=1,
  legend style={
    at={(0.5,0.8)},
    anchor=west,
    cells={anchor=west},
  },
]

\addplot+[dashed,
  color=red,
  mark=triangle,
  mark size=3pt,
  nodes near coords={ \pgfmathprintnumber[fixed,fixed zerofill,precision=3]{\pgfplotspointmeta}\%},
  nodes near coords style={xshift=-18pt,yshift=-12pt,},
  error bars/.cd, y dir=both,y explicit,
]
coordinates {
(449.00000, 0.36150) +- (0, 0.01329)
(1123.00000, 0.27458) +- (0, 0.01322)
(2021.00000, 0.23093) +- (0, 0.00866)
};

\addplot+[
  color=red,
  mark=triangle,
  mark size=3pt,
  nodes near coords={ \pgfmathprintnumber[fixed,fixed zerofill,precision=3]{\pgfplotspointmeta}\%},
  nodes near coords style={xshift=-18pt,yshift=-12pt,},
  error bars/.cd, y dir=both,y explicit,
]
coordinates {
(28.00000, 0.56928) +- (0, 0.01939)
(57.00000, 0.50861) +- (0, 0.02509)
(144.00000, 0.41923) +- (0, 0.01935)
(259.00000, 0.38126) +- (0, 0.01591)
};

\addplot+[dashed,
  color=green,
  mark=square,
  mark size=3pt,
  nodes near coords={ \pgfmathprintnumber[fixed,fixed zerofill,precision=3]{\pgfplotspointmeta}\%},
  nodes near coords style={xshift=18pt,},
  error bars/.cd, y dir=both,y explicit,
]
coordinates {
(449.00000, 0.38426) +- (0, 0.01673)
(1123.00000, 0.29857) +- (0, 0.00847)
(2021.00000, 0.26065) +- (0, 0.00971)
};

\addplot+[
  color=green,
  mark=square,
  mark size=3pt,
  nodes near coords={ \pgfmathprintnumber[fixed,fixed zerofill,precision=3]{\pgfplotspointmeta}\%},
  nodes near coords style={xshift=18pt,},
  error bars/.cd, y dir=both,y explicit,
]
coordinates {
(28.00000, 0.56347) +- (0, 0.01645)
(57.00000, 0.50443) +- (0, 0.02584)
(144.00000, 0.42575) +- (0, 0.01354)
(259.00000, 0.38917) +- (0, 0.01473)
};

\legend{, Fine-tuned (full parameter), , Fine-tuned (only attention layers)};
\node[black] at (rel axis cs:0.25,0.4) {one sample for each shape};
\node[black] at (rel axis cs:0.65,0.08) {all samples for each shape};

\end{axis}
\end{tikzpicture}
\caption{Effect of parameter-efficient fine-tuning under different downstream data regimes}
\label{fig:pefinetunesize}
\end{figure}
\end{modifiedblock}

\section{Interactive design application: WebWing}

\begin{figure}[ht]
    \centering
    \includegraphics[width=1\linewidth]{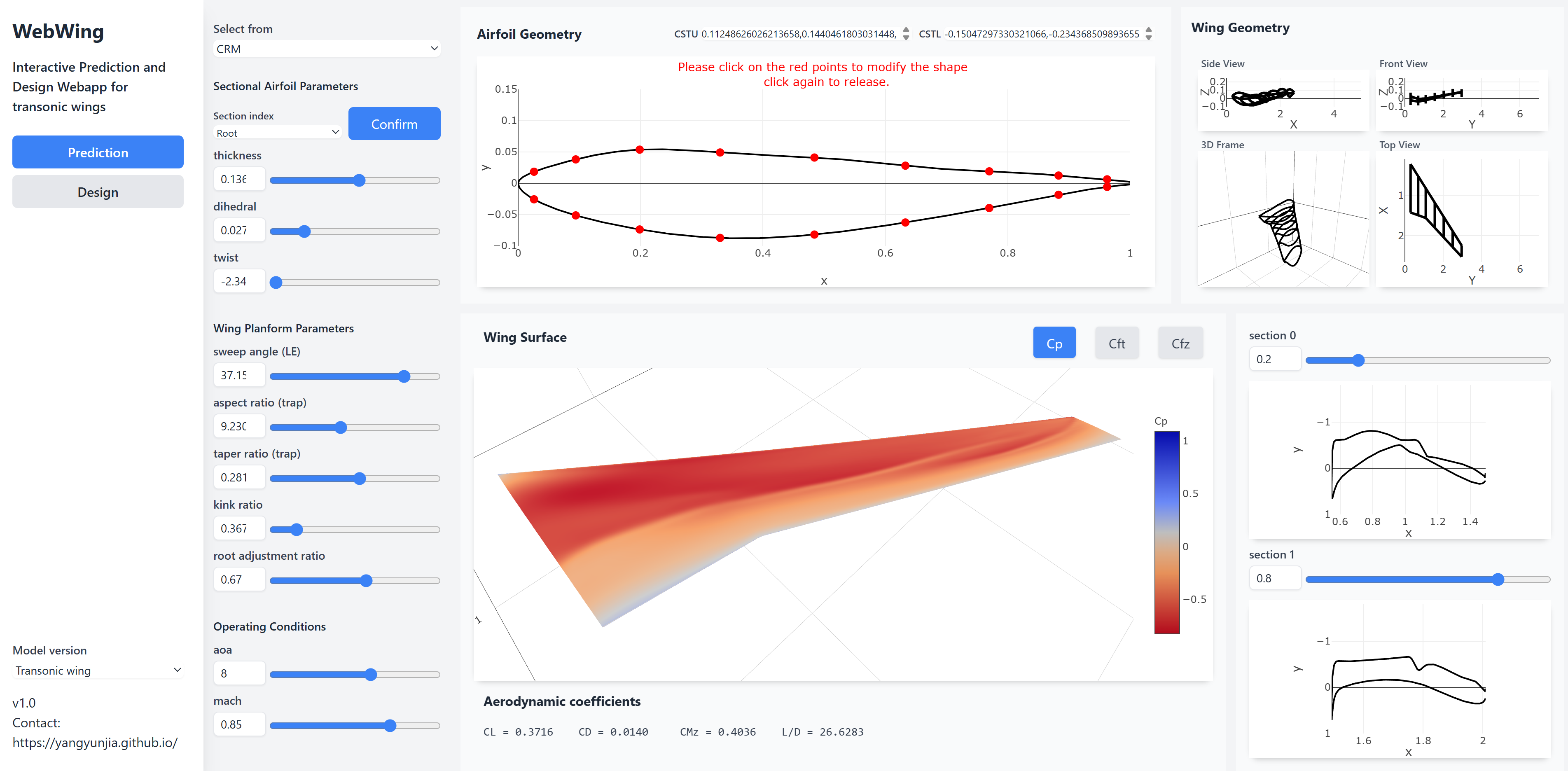}
    \caption{User interface for the interactive design tool Webwing, which remotely runs a pre-trained AeroTransformer model}
    \label{fig:webwing}
\end{figure}

Building on the strong generalization capability across diverse wing geometries, we developed and deployed an online interactive application, WebWing, to demonstrate the practical applicability of our pre-trained model for aerodynamic design. The app is accessible via a standard web browser at \href{https://webwing.pbs.cit.tum.de/}{https://webwing.pbs.cit.tum.de/} and allows users to explore wing aerodynamic behavior in real time. Figure \ref{fig:webwing} shows a screenshot of the application, which supports modification of airfoil shapes and local parameters (at up to seven control sections), global wing shape parameters, as well as operating conditions. These modifications can be achieved by uploading, inputting, or simply dragging the scroll bars and control points interactively. Upon each user modification, WebWing invokes the pre-trained L-size model to generate almost instant predictions for both surface flow fields (pressure and skin-friction distributions) and integrated aerodynamic coefficients. All results are visualized interactively within the interface, allowing users to inspect detailed distributions on selected sections. 

We believe that WebWing offers an intuitive, efficient tool for early-stage wing design, providing interactive aerodynamic feedback without the need for computationally intensive CFD simulations. By allowing designers to rapidly iterate over geometric variations and directly observe the associated aerodynamic responses, the application highlights the potential of pre-trained aerodynamic models in human-in-the-loop design workflows and conceptual aerodynamic analysis.

\section{Discussion}

\modified{Within this section, we discuss the proposed foundation-model paradigm in a broader perspective. The ultimate goal of establishing a foundation-style model is not intended for a single downstream task, but to serve as a \textit{common resource} that can be reused across multiple downstream applications, so that users do not need to repeatedly construct surrogate models from scratch for each new design problem. This justifies the large-scale investigation into dataset simulation and model training, and also motivates us to publicly release both the dataset and the trained model in this work. We envision that, in the long term, the community may benefit more from developing shared aerodynamic foundation models and adapting them to specific tasks.}

\modified{From this perspective, extending the proposed framework to more complex aerodynamic configurations is desirable. However, such an extension introduces several key challenges, which are discussed below.}

\paragraph{Geometric complexity}

\modified{Within the proposed framework, pre-training captures priors from simplified shapes, and fine-tuning adapts the model to a specific target configuration with greater detail. This works similarly for more complex configurations, such as the full aircraft, since there are methods to reduce the dimension of their design space for the pre-training stage. 
This enables the construction of pre-training datasets that capture the dominant geometric variations without exhaustively sampling all possible configurations. Results in this work support this idea by showing that a model pre-trained on such a dataset can be fine-tuned with limited data to restore effects introduced by configuration-specific details.} 

\paragraph{Computational cost}
\modified{The cost of generating the pre-training dataset can be viewed as a one-time investment and, more importantly, can be paralleled. Unlike solver-based optimization loops that iteratively call CFD, pre-training data generation can be fully parallelized on high-performance computing (HPC) systems. In this work, the pre-training dataset was generated over 4 months using only two HPC nodes. With a larger HPC, this time could be dramatically reduced. As a result, although the number of simulations is nontrivial, the process remains practically tractable due to its parallel nature.}

\paragraph{Model architecture}
\modified{The AeroTransformer can be extended to more complex geometries through more flexible input representations, such as projection-based or voxel-based embeddings, which allow handling configurations that cannot be mapped to structured two-dimensional grids. This further supports the applicability of the proposed framework beyond the current setting.}

\paragraph{}

\modified{Overall, while the present study serves as an initial step, the results indicate that shifting the learning process toward a reusable pre-training stage provides a viable and scalable pathway for surrogate modeling in increasingly complex aerodynamic design problems.}

\section{Conclusion}

This paper introduces a methodology that leverages large-scale pre-training to address the challenges of efficiently building an accurate surrogate model for predicting the aerodynamics of three-dimensional bodies. We systematically investigate both the benefits of pre-training and the strategies required for effective pre-training and fine-tuning. The main contributions are summarized as follows:

\begin{enumerate}
    \item We propose a two-stage methodology for building a local surrogate model for aerodynamic prediction. In the pre-training stage, a dataset generated from simplified shape parameterizations is used to reduce data-generation costs while ensuring broad coverage of the design space. Once trained, the model can be applied directly in a zero-shot manner or fine-tuned using a small number of high-fidelity samples within a task-specific design domain.
    
    \item We develop AeroTransformer, a Transformer based on the PDE-Transformer backbone, as the model architecture for the two-stage methodology. We tailor it to aerodynamic tasks by introducing operating-condition injection and an auxiliary aerodynamic loss. We train the AeroTransformer on the SuperWing dataset comprising nearly 30k samples, and the results show that it achieves a surface flow prediction error below 0.2\%. We further demonstrate that predicting surface flow as the primary output and integrating aerodynamic coefficients improves generalization, and the proposed extra loss helps balance the model errors. The model's scalability is also verified, as we found that both increased model capacity and larger pre-training datasets consistently enhance performance.

    \item We validate the effectiveness of the two-stage methodology by applying the pre-trained model to the surface flow prediction task of wings perturbed from the CRM. Compared to training a model from scratch, pre-training reduces prediction error by 84.2\%, with 450 task-specific samples. Even in zero-shot mode, the model can achieve good results and capture key flow features. We also demonstrate that improved pre-training with a larger model and larger dataset directly benefits downstream performance. Parameter-efficient fine-tuning techniques are also tested, further reducing computational cost and showing promising results, especially in a very small fine-tuning dataset.
    
\end{enumerate}

Although the current model and dataset remain modest relative to large language models, our results provide evidence that the foundation-model paradigm offers a promising path toward scalable and accurate surrogate modeling for three-dimensional aerodynamic design. 

\section*{Acknowledgments}

The authors gratefully acknowledge the financial support of the BMW Group. The authors would also like to thank Qiang Liu for his helpful comments. In this work, AI is used to assist coding for the WebWing App and to polish the expressions of this manuscript.

\section*{Data Availability}

The SuperWing dataset is available at \href{https://huggingface.co/datasets/yunplus/SuperWing}{https://huggingface.co/datasets/yunplus/SuperWing}, and the CRM-specific dataset is available at \href{https://huggingface.co/datasets/thuerey-group/CRMpert}{https://huggingface.co/datasets/thuerey-group/CRMpert}. The pre-trained AeroTransformer models (all three sizes) are published at \href{https://huggingface.co/thuerey-group/AeroTransformer}{https://huggingface.co/thuerey-group/AeroTransformer}. The implementation dependencies are summarized in  \href{https://github.com/tum-pbs/AeroTransformer}{https://github.com/tum-pbs/AeroTransformer}.

\bibliographystyle{unsrt}
\bibliography{sample}


\appendix

\renewcommand{\thesection}{\Alph{section}}
\titleformat{\section}
  {\large\bfseries\singlespacing\centering}
  {Appendix~\thesection.}{0.5em}{}

\renewcommand{\thesubsection}{\arabic{subsection}}
\titleformat{\subsection}
  {\normalsize\bfseries\singlespacing}
  {\thesubsection.}{0.5em}{}
\renewcommand{\thefigure}{\thesection\arabic{figure}}
\renewcommand{\thetable}{\thesection\arabic{table}}
\renewcommand{\theequation}{\thesection\arabic{equation}}
\setcounter{figure}{0}
\setcounter{table}{0}
\setcounter{equation}{0}

\section{Dimensionality reduction analysis of the wing shapes}\label{app:dim}

\modified{In the proposed framework, the pre-training dataset is designed to cover a broad range of wing geometries with moderate geometric fidelity, while the fine-tuning dataset focuses on a localized design space with richer geometric details. To better quantify the geometric complexity and the relationship between these datasets, we perform dimensionality reduction analyses to both of them. These analyses are intended to provide additional insight into the complexity of the geometric variations and the relationship between the two datasets.}

\subsection{PCA}

\modified{PCA provides a linear estimate of the intrinsic dimensionality of the dataset by analyzing the variance captured by orthogonal modes. We perform PCA on the flattened grid points of the wing shapes. Before applying PCA, all parameters are standardized to zero mean and unit variance to ensure that parameters with different physical scales contribute equally to the analysis.

PCA is then performed by computing the eigen-decomposition of the covariance matrix of the standardized data. Table~}\ref{tab:pca_modes} \modified{reports the cumulative explained variance ratios that estimate the number of modes required to represent a given fraction of the total geometric variation.}

\begin{table}[ht]
\centering
\caption{Number of PCA modes required to capture the variance of the geometry datasets.}
\label{tab:pca_modes}
\begin{tabular}{lcc}
\hline
Dataset & 99\% Energy & 99.9\% Energy \\
\hline
Pre-training dataset (SuperWing) & 5 & 11 \\
Fine-tuning dataset (CRM) & 11 & 42 \\
\hline
\end{tabular}
\end{table}

\modified{These results indicate that the fine-tuning dataset requires significantly more modes to represent the geometric variations, reflecting the richer local geometric details introduced in the fine-tuning stage. This observation is also consistent with the larger number of degrees of freedom used in the fine-tuning parameterization.

Nevertheless, it's worthy noting that in the pre-training dataset, the dominant modes primarily capture global planform variations such as aspect ratio, sweep, twist, and dihedral because these variations contribute relatively large geometric energy. In contrast, local variations such as the shapes of sectional airfoils typically contribute smaller geometric variance and may therefore be partially masked. For this reason, PCA is used here only as a complementary indicator rather than a definitive measure of geometric richness.}

\subsection{Visualization with $t$-distributed Stochastic Neighbor Embedding}

\modified{To further examine the relationship between the pre-training and fine-tuning datasets, we employ $t$-distributed Stochastic Neighbor Embedding ($t$-SNE)} \cite{van2008visualizing}\modified{, a nonlinear dimensionality reduction method that is particularly effective at preserving local neighborhood structure in high-dimensional data.}
\modified{We construct two $t$-SNE embeddings with a perplexity of 50 for the full pre-training and fine-tuning datasets, respectively. To enable a direct comparison, the other dataset is projected into the same low-dimensional space via $k$-nearest neighbors ($k$-NN) interpolation. 

The resulting visualization is shown in} Fig.~\ref{fig:tsne}. \modified{In subfigure (a), where the fine-tuning dataset is projected onto the pre-training embedding, the pre-training dataset exhibits a broad distribution, reflecting its global geometric diversity. In contrast, the fine-tuning dataset occupies a compact and localized region within this space. A zoomed-in view of this region is shown in the inset, which shows that the fine-tuning samples are densely clustered within a local neighborhood, while only a small number of pre-training samples are present in the same region. On the other hand, in subfigure (b), where the pre-training dataset is projected onto the fine-tuning embedding, the pre-training samples occupy only a subset of the space spanned by the fine-tuning dataset. This suggests that the fine-tuning dataset provides a more locally refined and densely sampled representation of the geometry space, with additional geometric variations that are not fully covered by the pre-training manifold.}

\begin{modifiedblock}
\begin{figure}[H]
    \centering
    \begin{subfigure}{0.48\textwidth}
        \includegraphics[width=\linewidth]{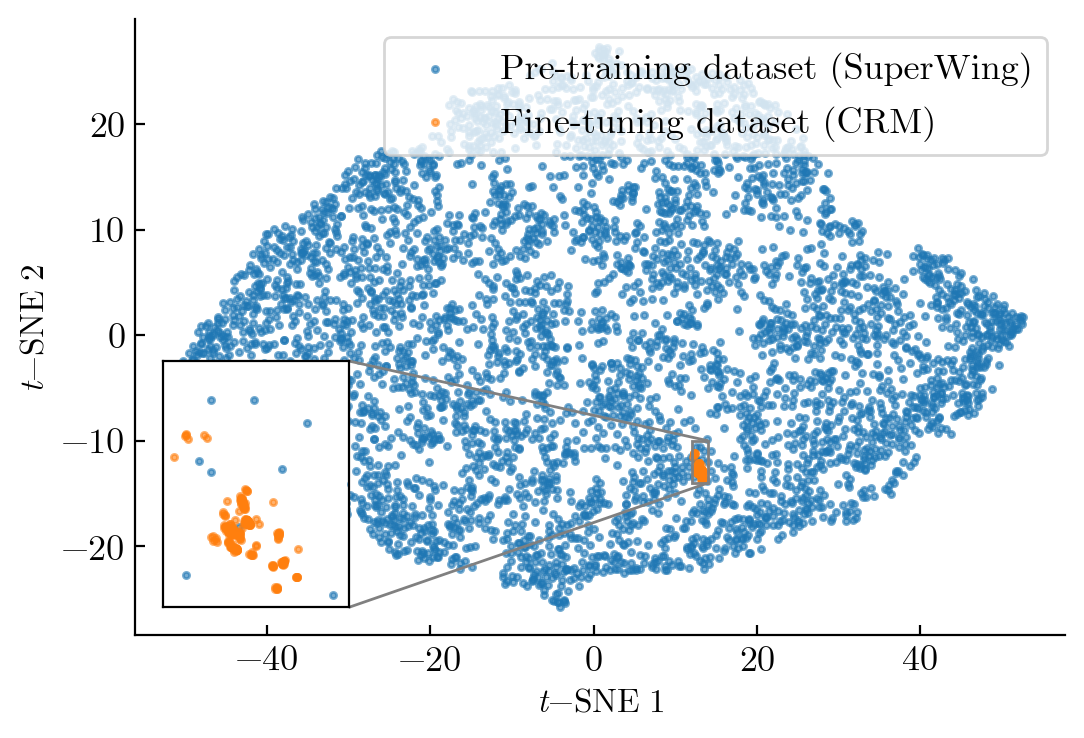}
        \caption{Projection of the CRM dataset to SuperWing}
    \end{subfigure}
    \hfill
    \begin{subfigure}{0.48\textwidth}
        \includegraphics[width=\linewidth]{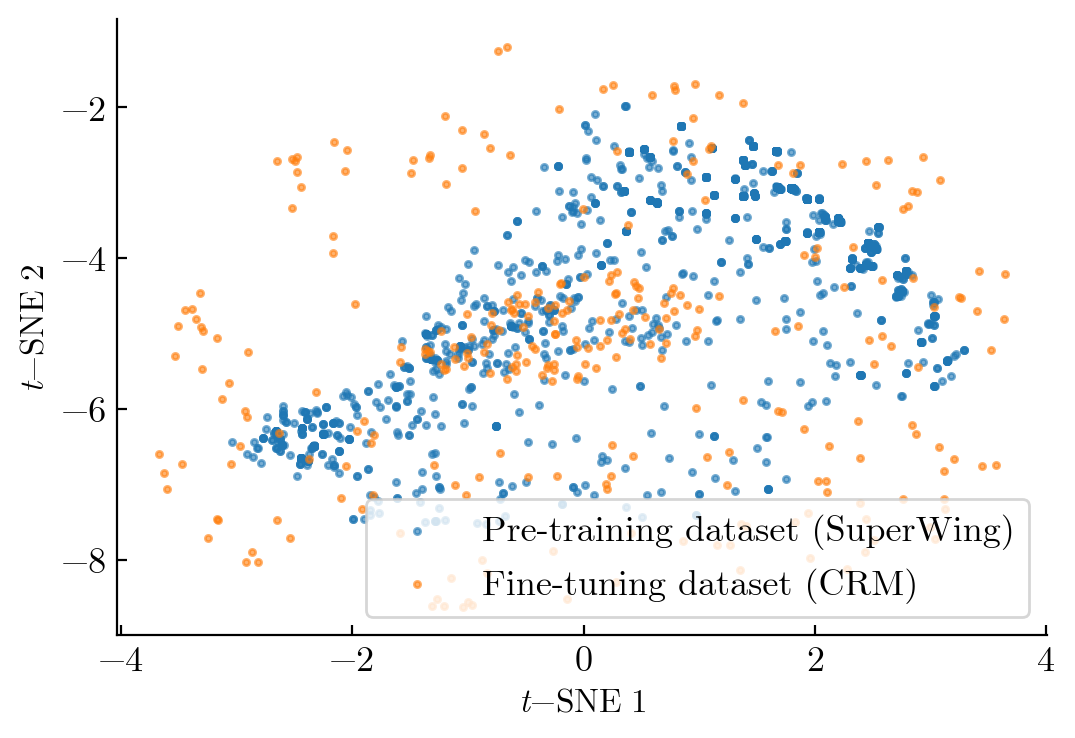}
    \caption{Projection of SuperWing to the CRM dataset}
    \end{subfigure}
    \caption{$t$-SNE visualization of the pre-training and fine-tuning dataset}
    \label{fig:tsne}
\end{figure}
\end{modifiedblock}


\section{Baseline models implementations}\label{app:baseline}
\setcounter{figure}{0}
\setcounter{table}{0}
\setcounter{equation}{0}

\modified{This appendix summarizes the implementations of the baseline models. We note that many model hyperparameters are essentially trade-offs between computational cost and performance; for example, a larger hidden dimension and a smaller patch size lead to better performance but also incur higher training costs. Considering this, the principle in selecting them is to enable a fair comparison while keeping computational cost reasonable. }

\subsection{Baseline models for prediction of surface flow}

\modified{We use the U-Net, ViT, and the Transolver for the baseline of surface-flow prediction. Considering the principle above, we control all model capacity by adjusting the hidden dimension $N_{\mathrm{hidden}}$ such that the total number of trainable parameters is of the same order (approximately 1M) as the S-size AeroTransformer. }

\paragraph{U-Net} U-Net, built upon CNN layers, is one of the most widely adopted architectures for flow field prediction. In this study, we use a U-Net with a symmetric encoder–decoder structure, and we incorporate operating conditions into the model by concatenating them with the latent representation. Specifically, the network is designed such that feature map operations occur along the airfoil circumferential direction, while the spanwise dimension remains unchanged. Both the encoder and decoder consist of six ResNet groups, each including a residual block for down- or up-sampling that halves or doubles the airfoil-circumferential resolution, followed by a standard residual block. The decoder concludes with a final convolutional layer that reduces the hidden channels to three output ones. \modified{The selection of the hyperparameters above is based on our previous experiment of wing flow field prediction on single-segment wings.}\cite{yang_rapid_2025}. \modified{In the present study, we modified the number of hidden dimensions in each block to achieve the proper size, which leads to $(8, 16, 26, 32, 32, 64)$ in each encoder block, and symmetrically in the decoder.}

\paragraph{Vision Transformer and Transolver} 

\modified{We introduced ViT in Sec.} \ref{sec:transformer}, \modified{and Transolver is one of the most successful extensions of the Transformer architecture to point clouds and unstructured meshes. It} retains the overall Transformer structure but selects a more physically-consistent tokenization strategy to establish the physics-embedded attention. The entire input field is not sliced into patches; instead, it is projected into a low-dimensional embedding space via learnable weighted projections. The self-attention is then performed in the projected low-dimensional space, and afterward, the field is reconstructed using the new tokens and the original weights. 

\modified{For Transformer-based models, we adopt a shared set of backbone configurations following common practices in the literature} \cite{wu_transolver_2024}. In particular, the number of layers $N_{\mathrm{layer}}=5$,  the number of attention heads  $N_\mathrm{heads} = 8$, and the MLP ratio $r_\mathrm{MLP} = 4$. \modified{In this unified setting, we only vary components intrinsic to each architecture, such as the patch partitioning strategy and positional encoding in ViT, or the design of the slicing operator $\mathcal{F}(\cdot)$ in Transolver. We conducted experiments on a portion of the SuperWing dataset, whose results were reported in our previous study} \cite{yang_eucass_2025}. 

For ViT, this leads us to use a uniform patch size of $p_H = p_W = 4$, yielding 2048 tokens, and a learnable matrix of shape $M \times N_\mathrm{hidden}$ for the positional embedding. For Transolver, the number of tokens is 64, and a $3 \times 3$ CNN with a stride of 1 and a padding of 1 is used to compute the weights for each slice, since the prediction task is based on a structural mesh. \modified{Again, the hidden dimension $N_{\mathrm{hidden}}$ was tuned to achieve a similar parameter count, which leads to 128 for both models.}

\subsection{Baseline models for prediction of aerodynamic coefficients}

Random Forest (RF) and Light Gradient Boosting Machine (LGBM) are used as two non-neural baselines for the geometry-to-performance prediction task. They directly predict aerodynamic coefficients from geometric parameters rather than from meshes. To ensure all inputs and outputs were on the same scale, we normalized the values using the 10th and 90th percentiles. 

\modified{To achieve the same model size of around 1M, we vary the number of estimators. The other hyperparameters follow the common recommendation, and by further tuning several parameters, we found they have a limited impact on the relative performance.}

\paragraph{Random Forest} RF model follows the standard bootstrap-aggregated formulation in which a collection of decision trees is trained on resampled subsets of the data. At each split, a random subset of features is selected, and final predictions are obtained by averaging the trees. In our implementation, the RF contains multiple regression trees, and we set the algorithm without a tree depth constraint.

\paragraph{Light Gradient Boosting Machine} The LGBM model represents a more expressive gradient-boosting approach, where trees are added sequentially to minimize the residual error of the current ensemble. We employ a configuration with no tree depth constraints, a learning rate of 0.05, and subsampling ratios of 0.8 for rows and 0.9 for features. 

\section{Pre-training configurations}\label{app:trainconfig}
\setcounter{figure}{0}
\setcounter{table}{0}
\setcounter{equation}{0}

\subsection{Total training steps}\label{app:steps}

To assess the effect of training steps on the performance of the pre-trained model, we conducted a pre-experiment using the L-size AeroTransformer on the full pre-training dataset. Models are trained with different numbers of optimization steps, ranging from 36.6k to 585.6k, while all other training settings remain identical. 

Figure \ref{fig:steps} presents the corresponding errors in surface flow prediction for the training and testing datasets, where we observe a consistent decrease in error as the number of training steps increases, indicating that the model continues to benefit from extended optimization and does not saturate within the tested range. Plus, the gap between training and testing errors remains moderate and stable across all step counts, suggesting that overfitting is not a primary concern in this pre-training regime. Although large training steps may further improve performance, we chose 585.6k for the main experiments, given the training budget.

\begin{modifiedblock}
\begin{figure}[H]
\centering
\small
\begin{tikzpicture}
\begin{axis}[
  width=0.6\textwidth,height=0.3\textwidth,
  axis lines=left,
  xmin=25,xmax=600,ymin=0.06,ymax=0.25,
  xmode=log,
  xlabel={Amount of pre-training steps},
  ylabel={Surface flow error ($SFE$)},
  xtick={36.6, 146.4, 292.8, 585.6},
  xticklabels={36.6k, 146.4k, 292.8k, 585.6k},
  yticklabel={
        \pgfmathparse{\tick}
        \pgfmathprintnumber[fixed,fixed zerofill,precision=2]{\pgfmathresult}\%
    },
  legend columns=1,
  legend style={
    at={(0.5,0.9)},
    anchor=west,
    draw=none
  },
]

\addplot+[
  color=black,
  dashed,
  mark=square,
  mark size=3pt,
  mark options={solid, fill=none},
]
coordinates {
(36.6, 0.16383)
(146.4, 0.13390)
(292.8, 0.09727)
(585.6, 0.07263)
};

\addplot+[
  color=black,
  mark=square,
  mark size=3pt,
  nodes near coords={
    \pgfmathprintnumber[fixed,fixed zerofill,precision=3]{\pgfplotspointmeta}\%
  },
  nodes near coords style={
    xshift=12pt,
    },
]
coordinates {
(36.6, 0.21979)
(146.4, 0.17760)
(292.8, 0.13689)
(585.6, 0.11036)
};

\legend{Training, Testing}
\end{axis}
\end{tikzpicture}

\caption{Pre-training performance of L-size AeroTransformer with different number of training steps}\label{fig:steps}
\end{figure}
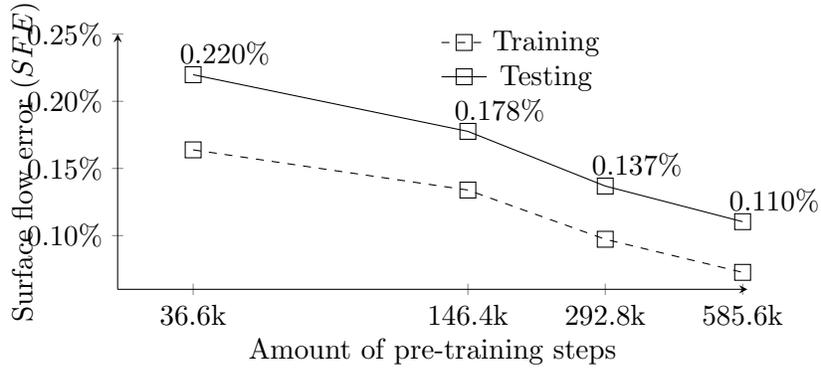
\end{modifiedblock}

\subsection{Gradient clipping}\label{app:clipping}

\begin{figure}[ht]
    \centering
    \includegraphics[width=0.5\linewidth]{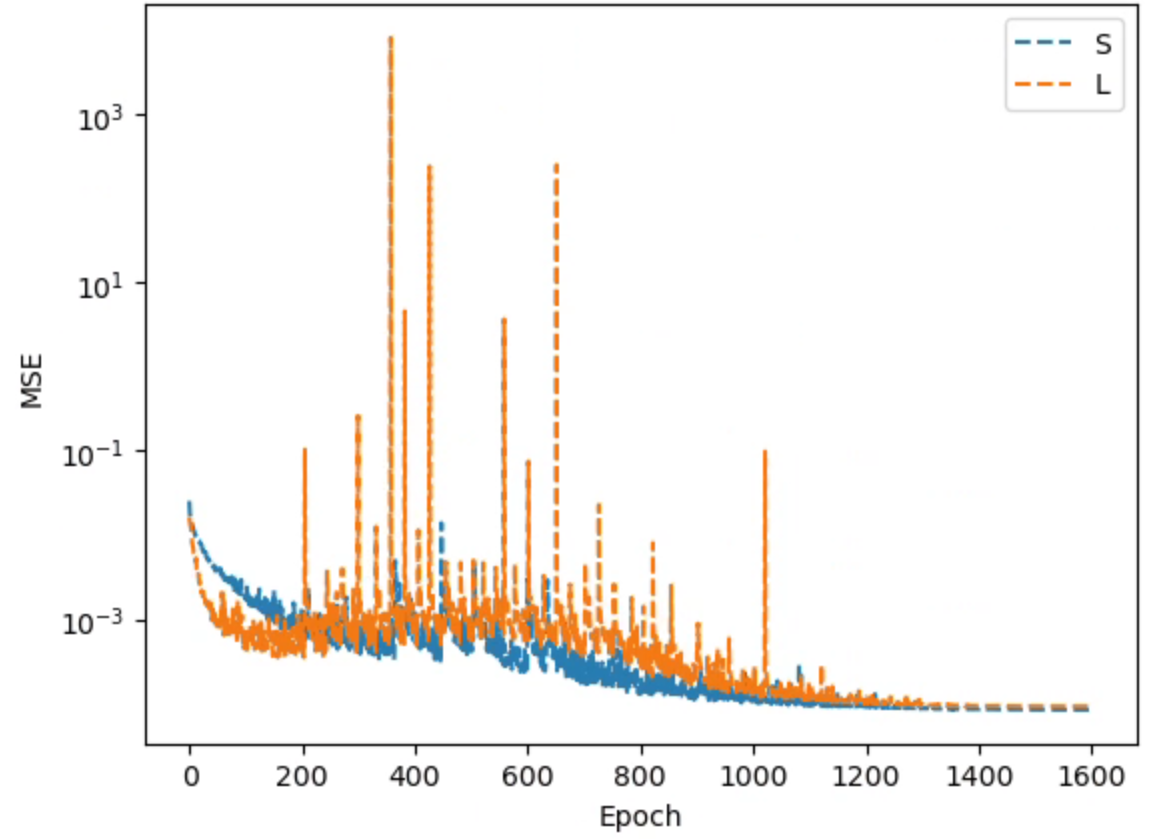}
    \caption{Training loss with norm clipping 1.0 for S and L model}
    \label{fig:lossSL}
\end{figure}

The stability of the gradient plays a critical role in training, especially for large models. In Fig. \ref{fig:lossSL}, we show the loss on the validation samples during the training of an S-size and L-size AeroTransformer with a full pre-training dataset. The two models are trained with the same settings, i.e., a maximum learning rate of $10^{-3}$, the default one-cycle scheduler, and gradient clipping with a norm of 1.0. Although this setting works well with the S-size model, it fails to stabilize the gradient on the L-size model, resulting in degraded performance. 

We tested several approaches to stabilize the gradient for the L-size model, including training the model with a lower learning rate and applying gradient clipping techniques, including norm clipping and exponential moving average (EMA) clipping. Table \ref{tab:clippingerrors} presents the prediction errors on the testing dataset using the aforementioned approaches, where norm clipping is effective, whereas a lower learning rate and EMA clipping fail to improve performance. Among the three norm value selections, the medium one, at 0.5, yields the best performance and is therefore used in the main experiments.

\begin{table}[htbp]
    \centering
    \caption{\label{tab:clippingerrors}Model performance comparison with and without aerodynamic loss term}
    \begin{tabular}{ccccccccc}
        \hline\hline
        \multicolumn{2}{c}{\multirow{2}{*}{Approaches}} & \multirow{2}{*}{\makecell{Max. \\ learning \\ rate}} & \multicolumn{6}{c}{Errors} \\ 
        &&& $\delta C_p$ & $\delta C_{f,\tau}$ & $\delta C_{f,z}$ & $\delta C_L$ & $\delta C_D$ & $\delta C_{M,z}$ \\
        &&& \multicolumn{3}{c}{\mdf Percentage relative error (\%)} & \mdf ($\times 10^{-3}$) & \mdf ($\times 10^{-4}$) & \mdf ($\times 10^{-3}$) \\
        \midrule
        \multicolumn{2}{c}{\multirow{2}{*}{w/o clipping}} & $10^{-3}$ & 0.180 & 0.137 & 0.165 &  1.56 & 1.34 & 1.82 \\ 
        & & $10^{-4}$ & 0.193 & 0.137 & 0.157 &  2.83 & 2.19 & 3.51 \\ 
        \multirow{3}{*}{\makecell{Norm \\ clipping}} & norm=1.0 & \multirow{3}{*}{$10^{-3}$} & 0.136 & 0.104 & 0.125 &  1.24 & 1.04 & 1.45  \\
        & norm=0.5 & & \textbf{0.123} & \textbf{0.095} & \textbf{0.113} &  \textbf{1.13} & \textbf{0.98} & \textbf{1.35}  \\
        & norm=0.2 & & 0.129 & 0.099 & 0.115 &  1.46 & 1.17 & 1.80 \\
        \multicolumn{2}{c}{EMA clipping} & $10^{-3}$ & 0.245 & 0.169 & 0.188 &  4.32 & 3.15 & 5.04 \\
        \hline\hline
    \end{tabular}
\end{table}

\section{Relationship between surface flow errors and aerodynamic coefficient errors}\label{app:mismatch}
\setcounter{figure}{0}
\setcounter{table}{0}
\setcounter{equation}{0}

\modified{This appendix provides evidence to better understand why lower pointwise surface flow error does not always lead to lower aerodynamic-coefficient error. We use the models with and without the aerodynamic loss term in Table}~\ref{tab:aeroloss} \modified{as a demonstration here. In particular, the model without aerodynamic loss ($\lambda=0$) achieves better pointwise surface flow accuracy, while the model with aerodynamic loss ($\lambda=0.5$) yields more accurate aerodynamic coefficients. We note that the analysis below could also be used to explain results from other parts.

As explained in Sec.} \ref{sec:losses}, \modified{one can decompose the coefficient error into the sum of local weighted contributions over the wing surface. Unlike the surface flow metric, which uniformly averages point-wise errors, the integrated coefficient depends on the \textit{signed} and \textit{spatially weighted} accumulation of these errors. Both of them lead to the discrepancy.}

\paragraph{Error distribution}

\modified{We first examine the spatial distribution of the pressure coefficient error $\delta C_{p,i}$. and its contribution to the lift coefficient $C_L$. Two representative samples are shown in Fig.}~\ref{fig:errordist}. \modified{The left two columns show the point-wise error $\delta C_{p,i}$. The regions with large error are located on the upper surface and near the shock wave in both models, and the model that predicts better surface fields generally exhibits smaller errors. Regarding the spanwise distribution of the error, we observe that the model that predicts better surface fields has larger error near the root, whereas the model that predicts better coefficients has larger error near the tip. Since the root region has a larger chord length, errors in this region contribute more during integration in the direction of lift. We demonstrate this by displaying the weighted error of $C_{p, i}$, where the pointwise error is multiplied by the sensitivity of the lift coefficient with respect to pressure, $g_{i} = \frac{\partial C_L}{\partial C_{p,i}}$. This sensitivity shows how much each local error contributes to the global integrated error to the first order. After weighting, a different pattern emerges where the errors near the wing root are amplified, while the errors near the tip are relatively suppressed. As a result, the model without aerodynamic loss exhibits larger weighted errors in the root region, despite having smaller point-wise error. }

\begin{modifiedblock}
\begin{figure}[H]
    \centering
    \includegraphics[width=1\linewidth]{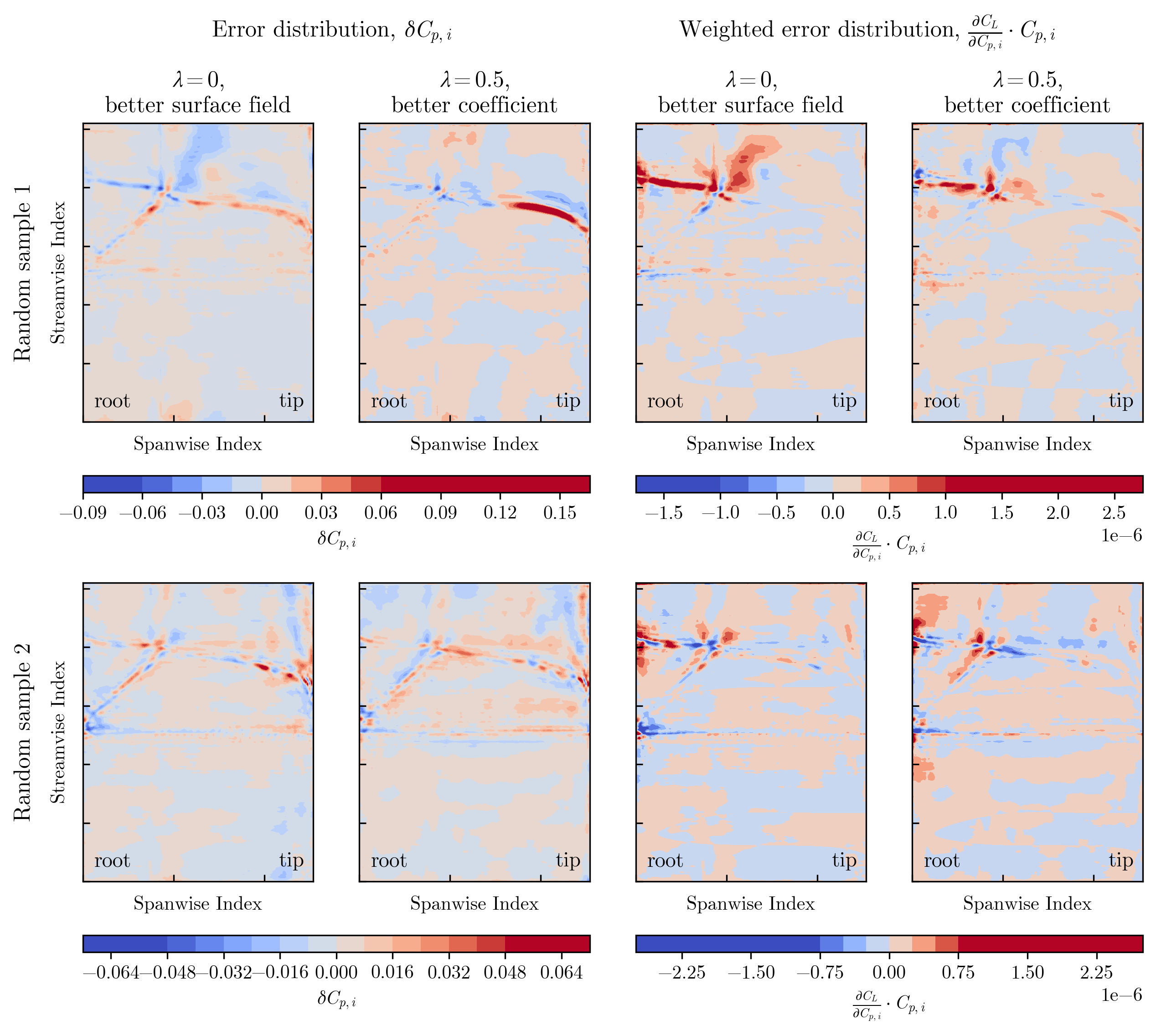}
    \caption{Distributions of point-wise errors $\delta C_p$, and the errors weighted by their sensitivity to the lift coefficients $g_{i} = \frac{\partial C_L}{\partial C_{p,i}}$}
    \label{fig:errordist}
\end{figure}
\end{modifiedblock}

\paragraph{Error offset}

\modified{We further analyze the relationship between surface flow error and drag coefficient error. Unlike lift, the contribution of point-wise error to drag is less dominated by spanwise geometric scaling. 

Figure}~\ref{fig:cperrordist} \modified{shows the distribution of weighted surface flow errors contributing to the drag coefficient. Although the model without aerodynamic loss exhibits a smaller overall error magnitude, the distribution is skewed, leading to a larger signed error accumulation. In contrast, the model with aerodynamic loss produces a more symmetric distribution, resulting in an offset between positive and negative contributions.}

\begin{modifiedblock}
\begin{figure}[H]
    \centering
    \includegraphics[width=0.5\linewidth]{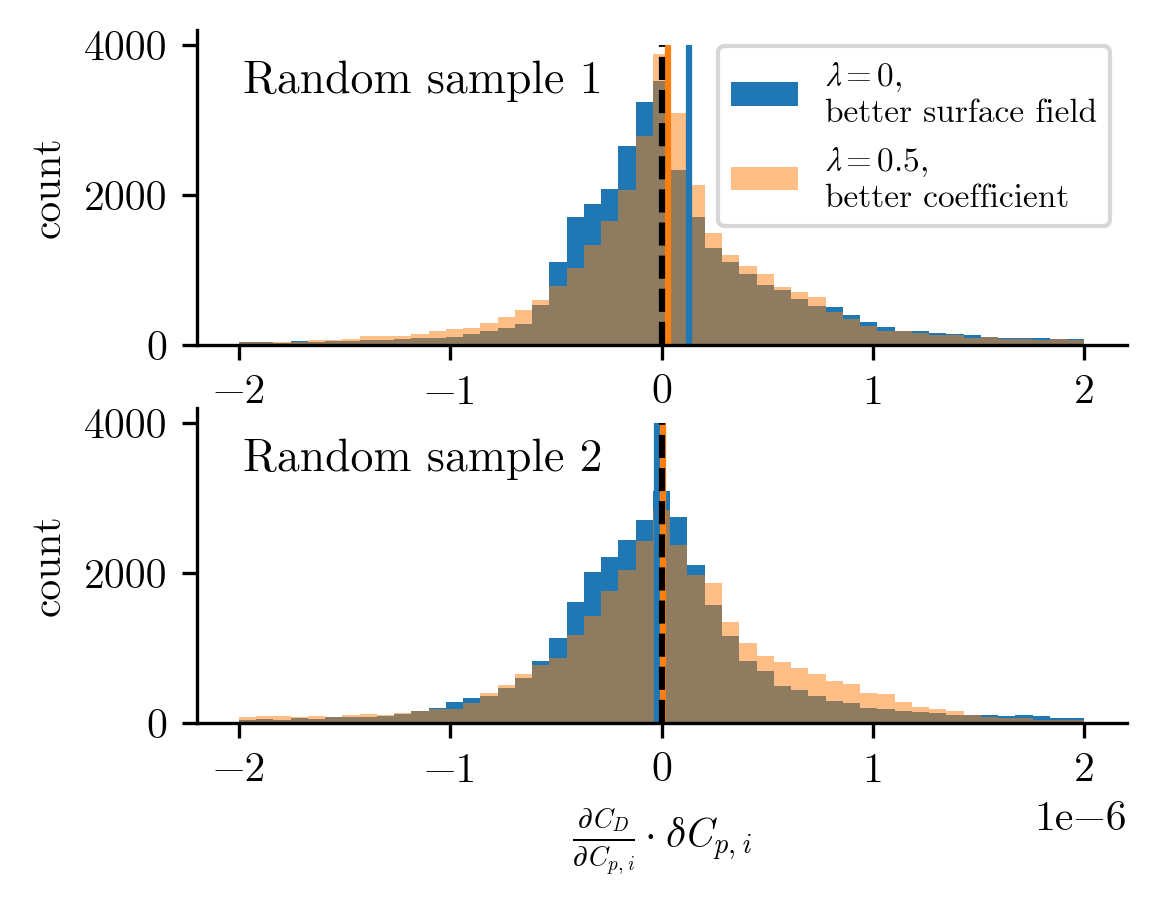}
    \caption{Histogram of point-wise error $\delta C_p$ weighted by their sensitivity to the drag coefficients $g_{i} = \frac{\partial C_D}{\partial C_{p,i}}$ (vertical lines showing the mean values of the corresponding histograms)}
    \label{fig:cperrordist}
\end{figure}
\end{modifiedblock}

\paragraph{}

\modified{The above analysis shows that minimizing pointwise surface flow error does not necessarily yield optimal predictions of aerodynamic coefficients. Instead, the aerodynamic loss term helps reshape the spatial distribution of errors, encouraging patterns that are more consistent with the underlying integral quantities, which, to some extent, are physically meaningful error distributions. This also provides an explanation for the improved generalization behavior observed when incorporating aerodynamic loss.}

\end{document}